\documentclass[Afour,sageh,times]{sagej}
\def\journalname{The International Journal of Robotics Research}
\usepackage{graphicx}
\usepackage{amsmath,amssymb}
\usepackage{array,tabularx,multirow,booktabs}
\usepackage{xcolor}
\usepackage{xurl}
\usepackage[colorlinks,bookmarksopen,bookmarksnumbered,citecolor=blue,urlcolor=blue]{hyperref}
\makeatletter
\let\IJRR@original@natbib@link\NAT@hyper@
\def\NAT@hyper@#1{\ifx\@captype\@undefined\mbox{\IJRR@original@natbib@link{#1}}\else\IJRR@original@natbib@link{#1}\fi}
\makeatother
\let\cite\citep

\newcolumntype{Y}{>{\raggedright\arraybackslash}X}
\newcolumntype{P}[1]{>{\raggedright\arraybackslash}p{#1}}
\begin{document}
\runninghead{Zhou et al.}
\title{Vision-Based Tactile Intelligence for Robotics: Hardware Design, Data Regimes, and Embodied Manipulation}

\author{Peng Zhou\affilnum{1,3,*}, Jun Hu\affilnum{1,2,*}, Sihan Chen\affilnum{3},
Zeqing Zhang\affilnum{4}, Haofei Ma\affilnum{5}, Zhenyu Lu\affilnum{6},
Sichao Liu\affilnum{7}, Xueqian Wang\affilnum{2}, Pai Zheng\affilnum{5},
Xiang Li\affilnum{2}, Shan Luo\affilnum{8}, Jia Pan\affilnum{3,\ensuremath{\dagger}},
David Navarro-Alarcon\affilnum{5,\ensuremath{\dagger}}, Chenguang Yang\affilnum{5}, and
Michael Yu Wang\affilnum{1}}
\affiliation{\affilnum{1}Great Bay University; \affilnum{2}Tsinghua University; 
\affilnum{3}The University of Hong Kong; \affilnum{4}Nanyang Technological University; 
\affilnum{5}The Hong Kong Polytechnic University; \affilnum{6}South China University of Technology; 
\affilnum{7}KTH Royal Institute of Technology; \affilnum{8}King's College London\\
\affilnum{*}Peng Zhou and Jun Hu contributed equally and share first authorship.\\
\affilnum{\ensuremath{\dagger}}\textbf{Corresponding authors:}
Jia Pan (\href{mailto:jpan@cs.hku.hk}{jpan@cs.hku.hk}) and
David Navarro-Alarcon (\href{mailto:dnavar@polyu.edu.hk}{dnavar@polyu.edu.hk}).}

\begin{abstract}
Advancing embodied robotics requires touch sensing that captures local physical interaction beyond the reach of external cameras. Vision-based tactile sensors (VBTSs) image contact-induced elastomer deformation at high spatial resolution, pairing rich physical cues with direct compatibility for modern visual learning and control. Whereas prior surveys often examine sensor hardware, perception algorithms, simulation, or datasets in isolation, progress toward embodied tactile intelligence increasingly requires viewing VBTSs as integrated sensing-and-learning systems in which materials, optics, calibration, and learning co-evolve. This review synthesizes the physical and computational foundations of vision-based tactile sensing across the full robotics pipeline. It first outlines representative sensor architectures, optical transduction cues, contact-mechanics and geometric-reconstruction models, and the limitations of their idealized assumptions. Linking physical choices to signal formation, calibration requirements, and coupled trade-offs, hardware development is organized around six design goals: miniaturization, large-area coverage, spatial resolution, force measurement performance, temporal performance, and multimodal sensing. Alongside a unit-explicit comparison of reported data budgets, perception, simulation, datasets, and deployment are then structured into five data and learning regimes: sensor-specific measurement, task-oriented tactile sensing, tactile data augmentation, cross-sensor generalization, and embodied tactile intelligence encompassing tactile world models and vision-language-action policies. Finally, to guide the transition of VBTSs into durable, reusable components of embodied robotic systems, the survey distills five open priorities: sensor durability, multimodal alignment, hardware-agnostic transfer, reactive feedback, and scalable data.
\end{abstract}

\maketitle

\section{Introduction}
\label{sec:introduction}

A fundamental challenge in robotics is to understand the physical properties of objects and use this knowledge to guide action. External camera images alone may not reliably reveal these properties or the contact states that arise during interaction. Tactile sensing provides complementary local information about surface geometry, texture, and contact mechanics, supporting object understanding and feedback control~\cite{tactilesensinghumanshumanoids2010,tactilesensingdexterousrobothandsreview2015,luotactileroboticsoutlook2025}.

Current robotic systems use a range of tactile sensing technologies, including piezoresistive, capacitive~\cite{captacrobustcapacitivesensingdistributedforcemappingparallelroboticgrasping2026}, magnetic~\cite{reskinversatilereplaceablelastingtactileskins2022}, piezoelectric, and triboelectric sensing, as well as sensors using electrical impedance tomography~\cite{deepneuralnetworkbasedelectricalimpedancetomographicsensingmethodologylargearearobotictactilesensing2021}. Broader surveys of tactile sensing and flexible electronics discuss these non-vision-based sensing technologies~\cite{tactilesensinghumanshumanoids2010,reviewtactilesensingtechnologiesapplicationsbiomedicalengineering2012,tactilesensingdexterousrobothandsreview2015,tactilesensorsreview2024,recentadvanceschallengestactilesensingroboticsfundamentalsapplications2025,innovationsflexibletactilesensorsthreedimensionalforcedetectionconstructionsignaldecouplingapplications2025,advancementsmachinelearningassistedflexibleelectronicstechnologiesapplicationsfutureprospects2026}. The spatial resolution and signal richness of a tactile sensor determine the information available for robotic perception, learning, and control. In many implementations, limited spatial resolution and signal richness constrain the reconstruction of contact force distributions and the characterization of fine-scale surface geometry and texture.

Vision-based tactile sensors (VBTSs) offer an alternative approach by using cameras to image contact-induced elastomer deformation. Because a single camera array observes thousands to millions of pixels across a compliant interface without wiring individual sensing elements, VBTSs decouple spatial density from electrical routing complexity. The resulting tactile images support detailed surface-texture perception and dense contact-force-field estimation~\cite{evaluationvisionbasedtactilesensor2004a,gelsighthighresolutionrobottactilesensorsestimatinggeometryforce2017}. More importantly, the image-based outputs of VBTSs allow tactile sensing to benefit directly from developments in computer vision and deep learning, supporting representation learning, semantic recognition, multimodal fusion, and tactile-guided manipulation. At the same time, optical transduction introduces distinct physical constraints---including illumination and optical-path volume, elastomer wear and viscoelastic hysteresis, and unit-to-unit optical calibration sensitivity---that motivate joint hardware and algorithmic design.

Early VBTS research explored the use of cameras for tactile sensing in robotics~\cite{reconstructingshapedeformablemembraneimagedata2000a,evaluationvisionbasedtactilesensor2004a,visionbasedtactilesensorusingtransparentelasticfingertipdexteroushandling2007a}. Subsequent developments included retrographic sensing~\cite{retrographicsensingmeasurementsurfacetextureshape2009} and GelSight~\cite{gelsighthighresolutionrobottactilesensorsestimatinggeometryforce2017} for detailed surface reconstruction, TacTip~\cite{tactipfamilysoftopticaltactilesensors3dprintedbiomimeticmorphologies2018} for biomimetic sensing, and DIGIT~\cite{digitnoveldesignlowcostcompacthighresolutiontactilesensorapplicationinhandmanipulation2020} for compact integration into robot hands.

As VBTSs evolve to support embodied intelligence, progress requires closer coordination between hardware and computational methods~\cite{improvingforcereconstructionperformancevisionbasedtactilesensorsoptimizingelasticbody2023,visionbasedtactilesensordesignusingphysicallybasedrendering2025,thintactthinvisionbasedtactilesensorlenslessimaging2025}. Existing reviews cover parts of this development. Hardware surveys analyze sensor design and performance~\cite{designdevelopmentvisionbasedtactilesensors2021,hardwaretechnologyvisionbasedtactilesensorreview2022,visionbasedtactilesensingperformanceparametersdevicedesign2025}; others examine sensing methods and signal processing~\cite{classificationvisionbasedtactilesensorsreview2025,markerfeaturesmultimodalfusionreviewvisionbasedtactilesensordesigndevelopment2025,forcemeasurementtechnologyvisionbasedtactilesensor2025a,whenvisionmeetstouchcontemporaryreviewvisuotactilesensorssignalprocessingperspective2024}. Data-generation reviews and broader robotics perspectives connect touch to learning and deployment~\cite{tactiledatagenerationapplicationsbasedvisuotactilesensorsreview2025,luotactileroboticsoutlook2025}. Recent surveys also analyze how sensor design and system integration influence VBTS performance in robotic applications~\cite{surveyvisionbasedtactilesensorshardwarealgorithmapplicationfuturedirection2025,coevolutionvisionbasedtactilesensorsstructuredominateddesigntaskbasedsystemintegrationreview2026}. These contributions provide valuable foundations. However, emerging approaches that integrate tactile sensing with large language models, vision-language models, vision-language-action (VLA) models, and vision-tactile-language-action (VTLA) models require closer coordination between sensor hardware and software for embodied reasoning and action generation. The connections among hardware components, learning methods, simulation, datasets, and embodied intelligence therefore need closer examination to guide this integration.

To address this gap, this review examines VBTSs as integrated sensing-and-learning systems, connecting hardware design, tactile perception, simulation, datasets, and embodied intelligence in robotic applications. We organize the discussion around two complementary views. First, hardware design is examined through six goals: 1) \emph{miniaturization}; 2) \emph{large-area coverage}; 3) \emph{spatial resolution}; 4) \emph{force measurement performance}; 5) \emph{temporal performance}; and 6) \emph{multimodal sensing}. For each design goal, we examine how hardware choices affect sensor observations and how measurement quality can guide hardware improvements. This analysis clarifies how hardware and algorithms interact in pursuing each design goal and highlights the associated design trade-offs. Second, we distinguish five data and learning regimes: 1) \emph{sensor-specific measurement}; 2) \emph{task-oriented tactile sensing}; 3) \emph{tactile data augmentation}; 4) \emph{cross-sensor generalization}; and 5) \emph{embodied tactile intelligence}. For each regime, we examine how tactile data and computational methods are used, how they depend on sensor hardware, and the capabilities and limitations of the resulting approaches. Together, these regimes provide a framework for understanding how different tactile sensing and learning approaches relate and contribute to embodied intelligence. This review focuses on VBTS hardware, tactile perception, simulation, datasets, embodied intelligence, and robotic manipulation, while incorporating other tactile sensing technologies where relevant for comparison, hybrid sensing, or cross-modality transfer.

\begin{figure*}[t]
\centering
\includegraphics[width=\textwidth]{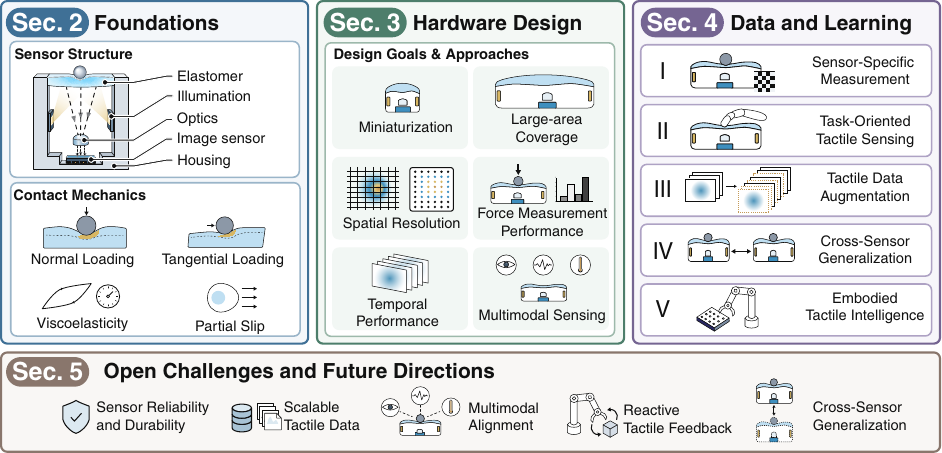}
\caption{Organization of the review and connections between its main topics. Section~\ref{sec:fundamentals} presents sensor structure, working principles, contact mechanics, and model limitations. Section~\ref{sec:hardware_taxonomy} organizes hardware design around six goals, while Section~\ref{sec:data_learning_regimes} distinguishes five data and learning regimes. Calibration and task results can inform hardware improvements. Section~\ref{sec:challenges} identifies five future research directions.}
\label{fig:survey_structure}
\end{figure*}

The remainder of this review is organized as follows. Section~\ref{sec:fundamentals} presents the structure and working principle of VBTSs and summarizes the theoretical foundations of contact mechanics and tactile reconstruction. Section~\ref{sec:hardware_taxonomy} examines hardware design through six goals. Section~\ref{sec:data_learning_regimes} synthesizes calibration, learning, simulation, datasets, and embodied intelligence through five regimes. Section~\ref{sec:challenges} discusses open challenges and future directions, and Section~\ref{sec:conclusion} concludes. Fig.~\ref{fig:survey_structure} summarizes the structure of this review and the connections among its main sections. An interactive companion website providing exploratory visualizations of the sensing principles, hardware taxonomy, audited data budgets, and future-directions roadmap is available at \url{https://gmi-lab-2025.github.io/VBTSwebsite/}.

\section{Mechanics and Mathematical Foundations of Vision-Based Tactile Sensing}
\label{sec:fundamentals}

\subsection{Sensor Structure and Working Principle}
\label{subsec:vbts_structure}

\begin{figure*}[t]
\centering
\includegraphics[width=\textwidth]{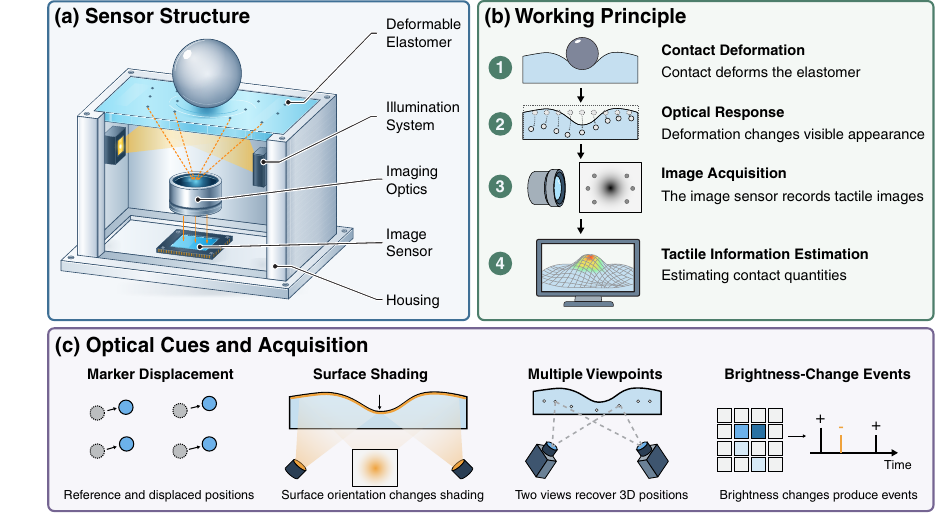}
\caption{Structure and working principle of a representative VBTS. (a) Four functional components: deformable elastomer, illumination system, imaging optics, and image sensor. (b) Contact deformation produces optical changes that are recorded as tactile images and interpreted using calibrated computational methods. (c) Marker displacement, surface shading, multiple viewpoints, and brightness-change events illustrate optical cues and acquisition approaches that can be combined.}
\label{fig:vbts_mechanics_pipeline}
\end{figure*}

A vision-based tactile sensor (VBTS) converts contact with a deformable interface into optical observations. Representative architectures contain four functional components, illustrated in Fig.~\ref{fig:vbts_mechanics_pipeline}(a). 1) The \emph{deformable elastomer} receives contact loads and converts them into surface deformation or internal motion; a reflective coating, markers, pins, or other visible features make this response observable. 2) The \emph{illumination system} lights the elastomer and its visible features, allowing the camera to capture changes caused by contact. 3) The \emph{imaging optics} guide and focus light from the elastomer onto the image sensor. 4) One or more \emph{image sensors} convert the focused light into tactile images, recording contact-induced changes in the elastomer's surface or visible features. Additional components, such as the housing and support structures, hold these elements in place.

The working principle of a VBTS can be described in four steps, as illustrated in Fig.~\ref{fig:vbts_mechanics_pipeline}(b). 1) \emph{Contact deformation.} When an object touches the sensor, the elastomer deforms according to the contact geometry and applied forces. 2) \emph{Optical response.} Under illumination, the deformation changes the appearance of the elastomer or its visible features. 3) \emph{Image acquisition.} The imaging optics direct light onto the image sensor, which records these changes as tactile images. 4) \emph{Tactile information estimation.} Computational methods analyze the images, often using a reference image or calibration, to estimate quantities such as contact geometry, deformation, or force.

In the second step, deformation produces optical cues that can be recorded using different imaging methods as shown in Fig.~\ref{fig:vbts_mechanics_pipeline}(c). 1) \emph{Marker displacement.} Markers on or within the elastomer move as it deforms. Their displacement relative to a reference image provides information about the deformation~\cite{evaluationvisionbasedtactilesensor2004a,tactipfamilysoftopticaltactilesensors3dprintedbiomimeticmorphologies2018}. 2) \emph{Surface shading.} A reflective coating is illuminated from different directions. Changes in the orientation of the deformed surface alter its shading, which can be used to reconstruct surface geometry~\cite{gelsighthighresolutionrobottactilesensorsestimatinggeometryforce2017,3dcalopensourcesoftwarelibrarydepthreconstructionvisionbasedtactilesensors2026}. 3) \emph{Multiple viewpoints.} Cameras observe the deformed surface or visible features from different directions. Matching the observations across views allows their three-dimensional positions to be estimated~\cite{gelstereobiotipselfcalibratingbionicfingertipvisuotactilesensorroboticmanipulation2024}. 4) \emph{Brightness-change events.} An event camera records local brightness changes as asynchronous events. These events can capture changes in visible features during dynamic contact~\cite{neurotacneuromorphicopticaltactilesensorappliedtexturerecognition2020,tactigraphasynchronousgraphneuralnetworkcontactanglepredictionusingneuromorphicvisionbasedtactilesensing2023}.

Raw tactile images and events contain information about contact, but they do not directly provide physical quantities such as deformation or force. Calibration relates these observations to known physical references. Different optical cues and desired tactile information require different calibration procedures. 1) \emph{Marker displacement calibration.} Spatial calibration relates distances in the image to distances on the sensor, allowing marker motion to be expressed in physical units~\cite{evaluationvisionbasedtactilesensor2004a}. Force and torque calibration relates marker displacement to known loads, allowing contact force and torque to be estimated~\cite{fingershapedgelforcesensormeasuringsurfacetractionfieldsrobotichand2010,effectiveestimationcontactforcetorquevisionbasedtactilesensorshelmholtzhodgedecomposition2019}. 2) \emph{Surface shading calibration.} Brightness-to-orientation calibration uses contact with an object of known shape to relate image brightness to surface orientation~\cite{gelsighthighresolutionrobottactilesensorsestimatinggeometryforce2017,3dcalopensourcesoftwarelibrarydepthreconstructionvisionbasedtactilesensors2026}. Force calibration relates changes in shading to known contact forces~\cite{utactunderwatervisionbasedtactilesensor2026}. 3) \emph{Multiple-viewpoint calibration.} Camera calibration determines the geometry of the imaging system; some designs also calibrate refraction as light passes through the sensor materials. These calibrations support three-dimensional reconstruction~\cite{zhang2023refractivegelstereo,gelstereobiotipselfcalibratingbionicfingertipvisuotactilesensorroboticmanipulation2024}. Force and torque calibration relates the reconstructed deformation to known loads~\cite{learningbasedsixaxisforcetorqueestimationgelstereo2022}. 4) \emph{Brightness-change event calibration.} Spatial calibration uses known contact positions to relate event locations to positions on the sensor~\cite{eventbasedoptotactileskin2026}. Force calibration relates marker displacement measured from events to known contact forces~\cite{ebtseventbasedtactilesensor2025}. Table~\ref{tab:optical_transduction_methods} summarizes the information, calibration requirements, and representative sensors associated with each optical cue.

\begin{table*}[t]
\centering
\caption{Optical sensing approaches in vision-based tactile sensors across the four primary optical cues introduced in Section~\ref{subsec:vbts_structure}.}
\label{tab:optical_transduction_methods}
\small
\renewcommand{\arraystretch}{1.15}
\setlength{\tabcolsep}{4pt}
\begin{tabularx}{\textwidth}{@{}P{0.15\textwidth}YYY@{}}
\toprule
\textbf{Observed cue} & \textbf{Extractable tactile information} & \textbf{Required calibration} & \textbf{Representative sensors} \\
\midrule
Marker displacement & 1) deformation; 2) contact position; 3) contact force; 4) contact torque; 5) slip & 1) spatial calibration for deformation and contact position~\cite{evaluationvisionbasedtactilesensor2004a}; 2) force and torque calibration~\cite{effectiveestimationcontactforcetorquevisionbasedtactilesensorshelmholtzhodgedecomposition2019} & GelForce~\cite{evaluationvisionbasedtactilesensor2004a}; TacTip~\cite{tactipfamilysoftopticaltactilesensors3dprintedbiomimeticmorphologies2018}; FingerVision~\cite{effectiveestimationcontactforcetorquevisionbasedtactilesensorshelmholtzhodgedecomposition2019}; GelSight~\cite{measurementshearslipgelsighttactilesensor2015} \\
Surface shading & 1) surface orientation; 2) contact geometry; 3) texture; 4) contact force & 1) brightness-to-orientation calibration~\cite{gelsighthighresolutionrobottactilesensorsestimatinggeometryforce2017}; 2) force calibration~\cite{utactunderwatervisionbasedtactilesensor2026} & GelSight~\cite{gelsighthighresolutionrobottactilesensorsestimatinggeometryforce2017}; DIGIT~\cite{3dcalopensourcesoftwarelibrarydepthreconstructionvisionbasedtactilesensors2026}; UTact~\cite{utactunderwatervisionbasedtactilesensor2026} \\
Multiple viewpoints & 1) 3D surface shape; 2) deformation; 3) contact position; 4) contact force; 5) contact torque & 1) camera and refraction calibration for 3D geometry~\cite{zhang2023refractivegelstereo}; 2) force and torque calibration~\cite{learningbasedsixaxisforcetorqueestimationgelstereo2022} & GelStereo BioTip~\cite{gelstereobiotipselfcalibratingbionicfingertipvisuotactilesensorroboticmanipulation2024}; GelStereo~\cite{learningbasedsixaxisforcetorqueestimationgelstereo2022} \\
Brightness-change events & 1) feature motion; 2) contact position; 3) contact angle; 4) contact force; 5) slip; 6) texture & 1) spatial calibration for contact position~\cite{eventbasedoptotactileskin2026}; 2) force calibration~\cite{ebtseventbasedtactilesensor2025} & NeuroTac~\cite{neurotacneuromorphicopticaltactilesensorappliedtexturerecognition2020}; Evetac~\cite{evetaceventbasedopticaltactilesensorroboticmanipulation2024} \\
\bottomrule
\end{tabularx}
\end{table*}

\subsection{Contact Mechanics and Tactile Reconstruction}
\label{subsec:normal_mechanics}

During contact, normal and tangential forces deform the elastomer, but a VBTS only observes this deformation through tactile images. Recovering physical contact quantities from these images therefore relies on two complementary foundations. First, contact-mechanics models relate applied loads, material properties, and contact geometry to elastomer deformation and local slip, providing the physical basis for inferring contact force, pressure distributions, torque, and incipient slip from observed deformation. Second, geometric reconstruction recovers the three-dimensional deformed contact surface directly from optical cues such as surface shading. The following models summarize these mechanical and geometric foundations.

\textbf{Hertz contact mechanics.} When an object presses against a VBTS, the elastomer indents and forms a contact area that grows with the normal load. Hertz contact mechanics models the relationship between applied normal force and elastic material deformation. For frictionless, non-adhesive contact between an elastic sphere and a thick elastic surface, it relates the contact radius $a$, indentation $\delta$, and normal load $F_n$~\cite{johnson1985contact}:
\begin{equation}
 a=\sqrt{R\delta},\qquad F_n=\frac{4}{3}E^*\sqrt{R}\,\delta^{3/2},
 \label{eq:hertz}
\end{equation}
where $R$ is the effective radius of curvature and $E^*$ is the effective elastic modulus given by
\begin{equation}
 \frac{1}{E^*}=\frac{1-\nu_1^2}{E_1}+\frac{1-\nu_2^2}{E_2}.
\end{equation}
Here $E_i$ and $\nu_i$ ($i=1,2$) are Young's modulus and Poisson's ratio of the two contacting bodies. These relations explain how contact area and indentation can provide information about the normal load. They also show why elastomer stiffness and contact geometry matter when selecting a sensing material and calibrating force estimates. For example, EasyCalib uses a Hertz-based contact model with corrections for sensor geometry to identify mechanical parameters for force reconstruction~\cite{easycalib2024}. However, these relations may not accurately describe force and deformation in practical tactile sensors. Finite pad thickness, rigid backing, adhesion, large strain, and non-spherical contacts change the relation.

\textbf{Linear viscoelasticity.} A viscoelastic elastomer can respond differently depending on how quickly it is pressed and how long it remains in contact. For example, after an object is removed, the elastomer may take time to return to its original shape even though the contact force has disappeared. For a one-dimensional linear-viscoelastic reference model with zero initial strain, the Boltzmann superposition principle relates the stress $\sigma(t)$ at time $t$ to the history of strain~\cite{husain2005viscoelasticity}:
\begin{equation}
 \sigma(t)=\int_0^t E_{\mathrm{rel}}(t-t')\dot{\varepsilon}(t')\,dt',
 \label{eq:relaxation}
\end{equation}
where $E_{\mathrm{rel}}$ is the relaxation modulus, $t'$ is the past-time integration variable, and $\dot{\varepsilon}(t')=d\varepsilon(t')/dt'$ is the strain rate (the time derivative of strain $\varepsilon$). This model shows how the material response depends on its deformation history. For force estimation, calibration should therefore consider loading, unloading, and sustained contact under the speeds and durations expected in use. For example, GelSight experiments showed that after a probe was moved sideways and held still, the contact force and marker positions continued to change over time~\cite{measurementshearslipgelsighttactilesensor2015}.

\textbf{Boussinesq solution.} A VBTS observes elastomer deformation, which can be used to estimate how contact pressure is distributed across the sensor surface. For a homogeneous, isotropic, linear-elastic half-space, the Boussinesq solution relates the normal pressure distribution $p_n$ to the resulting normal surface displacement $w$~\cite{johnson1985contact}:
\begin{equation}
 w(\mathbf{x})=\frac{1-\nu^2}{\pi E}
 \int_{\Omega}\frac{p_n(\boldsymbol{\xi})}
 {\|\mathbf{x}-\boldsymbol{\xi}\|}\,dA_{\boldsymbol{\xi}},
 \label{eq:boussinesq}
\end{equation}
where $\mathbf{x}$ is the observation point on the surface, $\boldsymbol{\xi}$ is a loaded point within the contact region $\Omega$, $\|\mathbf{x}-\boldsymbol{\xi}\|$ is the Euclidean distance between them, $dA_{\boldsymbol{\xi}}$ is the infinitesimal area element at $\boldsymbol{\xi}$, and $E$ and $\nu$ are Young's modulus and Poisson's ratio of the elastic medium. The integral smooths the pressure distribution. Recovering pressure from sampled deformation therefore requires both an appropriate boundary model and control of noise amplification. For example, GelForce uses Boussinesq and Cerruti elastic solutions to reconstruct contact force distributions from marker displacements~\cite{kamiyama2005traction}.

\textbf{Elastic Green tensor.} The elastic Green tensor extends the normal-force relationship described by the Boussinesq solution to include tangential forces and their combined effects on elastomer deformation. Under small-strain elasticity, the Green tensor relates the distribution of these forces to the resulting displacement~\cite{johnson1985contact}:
\begin{equation}
 u_i(\mathbf{x})=\sum_j\int_{\Omega}
 G_{ij}(\mathbf{x},\boldsymbol{\xi})p_j(\boldsymbol{\xi})\,dA_{\boldsymbol{\xi}},
 \label{eq:green}
\end{equation}
where $i,j\in\{x,y,z\}$ index the spatial coordinate directions, $u_i(\mathbf{x})$ is the displacement in direction $i$ at observation point $\mathbf{x}$, $p_j(\boldsymbol{\xi})$ is the contact traction in direction $j$ at loaded point $\boldsymbol{\xi}$ over the contact region $\Omega$ with area element $dA_{\boldsymbol{\xi}}$, and $G_{ij}(\mathbf{x},\boldsymbol{\xi})$ is the corresponding component of the elastic Green tensor. The response depends on the elastomer material and boundary conditions. Normal and tangential forces can both contribute to the observed displacement, so their effects must be considered when estimating contact forces~\cite{evaluationvisionbasedtactilesensor2004a,sferrazza2019ground}. For example, GelForce uses elastic response models to estimate the distribution of contact forces from marker motion~\cite{kamiyama2005traction}.

\textbf{Helmholtz--Hodge decomposition.} Marker motion across a VBTS surface forms a displacement field. Helmholtz--Hodge decomposition separates this field into components that describe different displacement patterns~\cite{effectiveestimationcontactforcetorquevisionbasedtactilesensorshelmholtzhodgedecomposition2019}:
\begin{equation}
 \mathbf{u}_{\parallel}=\nabla\phi+\nabla^{\perp}\psi+\mathbf{h},
 \label{eq:hhd}
\end{equation}
where $\mathbf{u}_{\parallel}$ is the in-plane (tangential) surface displacement field, $\phi$ and $\psi$ are scalar potentials whose gradient $\nabla\phi=(\partial_x\phi,\partial_y\phi)$ and rotated gradient $\nabla^{\perp}\psi=(-\partial_y\psi,\partial_x\psi)$ (with $\partial_x$ and $\partial_y$ denoting partial derivatives with respect to surface coordinates $x$ and $y$) represent the curl-free and rotational components, respectively, and $\mathbf{h}$ is a harmonic component. These components provide features for estimating contact force and torque. For example, a FingerVision-based method calibrates them against known loads to estimate contact forces and torque about the surface normal~\cite{effectiveestimationcontactforcetorquevisionbasedtactilesensorshelmholtzhodgedecomposition2019}.

\textbf{Cattaneo--Mindlin partial slip.} A VBTS can observe changes in surface or marker motion as an object begins to slip. The Cattaneo--Mindlin model divides an ideal circular contact into a central stick region, where the surfaces do not slide relative to each other, and an outer slip region, where they slide. Under these ideal conditions, increasing tangential force causes the central stick region to shrink and the surrounding slip region to expand. For elastically similar bodies under a constant normal load $F_n$, with friction coefficient $\mu$ and a monotonically increasing tangential load $F_t$, the radius $c$ of the central stick region is~\cite{johnson1985contact}:
\begin{equation}
 \frac{c}{a}=\left(1-\frac{F_t}{\mu F_n}\right)^{1/3},\qquad 0\le F_t\le\mu F_n,
 \label{eq:partial_slip}
\end{equation}
where $a$ is the contact radius. For a circular Hertz contact with peak pressure $p_0=3F_n/(2\pi a^2)$, the model gives the shear-traction magnitude $\tau(r)$ at radial distance $r$ from the contact center~\cite{johnson1985contact}:
\begin{equation}
 \frac{\tau(r)}{\mu p_0}=\begin{cases}
 \sqrt{1-\frac{r^2}{a^2}}-\dfrac{c}{a}\sqrt{1-\frac{r^2}{c^2}},&r\le c,\\[3pt]
 \sqrt{1-\frac{r^2}{a^2}},&c<r\le a.
 \end{cases}
\end{equation}
At $F_t=\mu F_n$, the stick region vanishes. These changes help explain why local motion can provide an early indication of slip. For example, GelSight studies use spatial differences in surface and marker motion to detect shear and incipient slip~\cite{measurementshearslipgelsighttactilesensor2015,improvedgelsighttactilesensormeasuringgeometryslip2017}.

\textbf{Photometric reconstruction.} For surface-shading sensors, the camera records changes in brightness. Photometric stereo estimates surface orientation from the calibrated optical response and uses it to recover shape~\cite{woodham1980photometric,gelsighthighresolutionrobottactilesensorsestimatinggeometryforce2017}. Let $z(x,y)$ describe the height of an observed surface over in-plane coordinates $(x,y)$, with surface gradients $g_x=\partial z/\partial x$ and $g_y=\partial z/\partial y$. Its unit surface normal is~\cite{retrographicsensingmeasurementsurfacetextureshape2009}:
\begin{equation}
 \mathbf{n}=\frac{[-g_x,-g_y,1]^{\mathsf T}}
 {\sqrt{1+g_x^2+g_y^2}},
\end{equation}
where $^{\mathsf T}$ denotes the transpose. Controlled illumination and a calibrated reflectance mapping can estimate these gradients from image intensity. Integrating noisy gradients in a least-squares sense gives a Poisson problem~\cite{gelsighthighresolutionrobottactilesensorsestimatinggeometryforce2017}:
\begin{equation}
 \nabla^2 z=\frac{\partial g_x}{\partial x}+\frac{\partial g_y}{\partial y},
 \label{eq:poisson}
\end{equation}
where $\nabla^2 z=\partial^2 z/\partial x^2+\partial^2 z/\partial y^2$ is the two-dimensional Laplacian of $z$, solved with specified boundary conditions or a height reference~\cite{3dcalopensourcesoftwarelibrarydepthreconstructionvisionbasedtactilesensors2026}. The reconstructed shape describes the sensor surface during contact. It does not directly provide contact force and may differ from the object's original shape if the object also deforms. 

Together, these seven models cover elastomer response, force--displacement relations, displacement-field decomposition, partial slip, and surface-shape reconstruction. Table~\ref{tab:mechanics_sensor_applications} compares what each model explains, its main limitation, and representative VBTS applications.

\begin{table*}[t]
\centering
\caption{Theoretical foundations of vision-based tactile sensing summarized from the seven models and methods in Section~\ref{subsec:normal_mechanics}.}
\label{tab:mechanics_sensor_applications}
\small
\renewcommand{\arraystretch}{1.15}
\setlength{\tabcolsep}{4pt}
\begin{tabularx}{\textwidth}{@{}P{0.19\textwidth}P{0.27\textwidth}P{0.23\textwidth}Y@{}}
\toprule
\textbf{Theory or method} & \textbf{What it explains} & \textbf{Limitation} & \textbf{Example applications} \\
\midrule
Hertz contact mechanics & The relationship between normal load, indentation, contact area, and material properties for ideal elastic contact. & Assumes ideal contact and a thick elastic material. & Tac3D: material calibration~\cite{easycalib2024} \\
Linear viscoelasticity & How deformation and stress depend on loading history and time. & The one-dimensional model cannot fully describe complex 3D deformation. & GelSight: time-dependent shear response~\cite{measurementshearslipgelsighttactilesensor2015} \\
Boussinesq solution & How normal pressure produces displacement in an elastic half-space. & Assumes a thick elastic material; pressure estimates can be sensitive to noise. & GelForce: contact-force reconstruction~\cite{kamiyama2005traction} \\
Elastic Green tensor & How normal and tangential forces produce coupled displacement fields. & Depends on accurate material and boundary models. & GelForce: normal and tangential force estimation~\cite{kamiyama2005traction} \\
Helmholtz--Hodge decomposition & How a displacement field can be represented by gradient and rotational components. & Does not uniquely identify force or torque. & FingerVision: force and torque estimation~\cite{effectiveestimationcontactforcetorquevisionbasedtactilesensorshelmholtzhodgedecomposition2019} \\
Cattaneo--Mindlin partial slip & How slip can develop around a central stick region under tangential loading. & Assumes ideal contact geometry and friction. & GelSight: incipient slip detection~\cite{measurementshearslipgelsighttactilesensor2015} \\
Photometric reconstruction & How surface orientation is estimated from calibrated shading and integrated to recover surface shape. & Sensitive to illumination and calibration errors. & GelSight: surface-shape reconstruction~\cite{gelsighthighresolutionrobottactilesensorsestimatinggeometryforce2017} \\
\bottomrule
\end{tabularx}
\end{table*}

\subsection{Limitations of Theoretical Models}
\label{subsec:coupled_inverse}

The analytical models in Section~\ref{subsec:normal_mechanics} clarify how contact loads, material deformation, and optical observations relate under idealized conditions, yet practical VBTSs diverge from these assumptions in four recurring ways.

\textbf{Finite geometry and boundary constraints.} Half-space formulations such as Hertz contact mechanics~(\ref{eq:hertz}) and the Boussinesq integral~(\ref{eq:boussinesq}) assume an infinitely thick, flat elastic medium under frictionless, non-adhesive, small-strain loading. Practical VBTS pads have finite thickness bonded to a rigid transparent window or curved shell, which constrains subsurface deformation and stiffens the effective contact response as indentation approaches a noticeable fraction of pad thickness. Curved fingertips, edges, protective coatings, surface adhesion, and non-spherical indenters further alter the contact-area and pressure-distribution relations. Likewise, ideal Cattaneo--Mindlin partial-slip behavior~(\ref{eq:partial_slip}) assumes circular Hertzian contact with uniform Coulomb friction, whereas real manipulation involves irregular contact patches, multi-point contact, compliant objects, and spatially varying surface roughness.

\textbf{Nonlinear and time-dependent material mechanics.} Soft silicone elastomers routinely undergo large strains and exhibit hyperelastic stiffening, stress relaxation, creep, and loading--unloading hysteresis that exceed one-dimensional linear viscoelasticity~(\ref{eq:relaxation}). Because elastomer stress depends on strain rate, hold duration, and prior loading history, a single static force--displacement lookup table can produce systematic bias during rapid tapping, sustained grasping, or cyclic sliding. Material properties also drift over time with temperature changes, mechanical fatigue, and surface wear.

\textbf{Coupled and ill-posed inverse mappings.} Recovering contact tractions from observed surface deformation requires inverting integral operators such as the Boussinesq~(\ref{eq:boussinesq}) and elastic Green-tensor~(\ref{eq:green}) relations. Because elastic media act as spatial low-pass filters, the forward mapping smooths high-frequency pressure variations; consequently, its inverse is ill-posed and amplifies small marker-tracking or depth-estimation errors unless stabilized through regularization, prior mechanical models, or dense finite-element representations. In addition, normal and tangential loads couple within the same three-dimensional displacement field, and kinematic decompositions such as Helmholtz--Hodge decomposition~(\ref{eq:hhd}) separate geometric flow patterns rather than uniquely determining physical force and torque without device-specific calibration or boundary-condition modeling.

\textbf{Optical and calibration sensitivity.} Photometric reconstruction~(\ref{eq:poisson}) depends on stable, well-characterized illumination and surface reflectance. Spatially varying lighting, specular highlights, grazing-angle shadowing, ambient light leakage, refraction through clear substrates, and coating degradation all introduce gradient errors that accumulate during Poisson integration into low-frequency depth drift. Moreover, geometric reconstruction recovers the deformed sensor surface rather than the undeformed object; when the contacted object is itself deformable, separating object compliance from elastomer deformation requires additional force, motion, or multimodal cues.

Taken together, these four gaps explain why closed-form analytical models rarely suffice on their own for accurate real-world tactile perception. Overcoming them requires coordinating physical sensor design---to make target contact cues optically observable and mechanically well-conditioned (Section~\ref{sec:hardware_taxonomy})---with calibrated, physics-informed, and data-driven learning methods (Section~\ref{sec:data_learning_regimes}).
\section{Hardware Design: Design Goals and Trade-offs}
\label{sec:hardware_taxonomy}

Translating the sensing principles in Section~\ref{sec:fundamentals} into practical vision-based tactile sensors requires choices across the elastomer, illumination, imaging optics, and image sensor. These hardware choices govern how contact is captured in tactile images, what physical information can be obtained, and what calibration is required to interpret them. Because different robotic tasks impose competing physical constraints, no single hardware design excels in all respects.

\begin{figure*}[t]
\centering
\includegraphics[width=\textwidth]{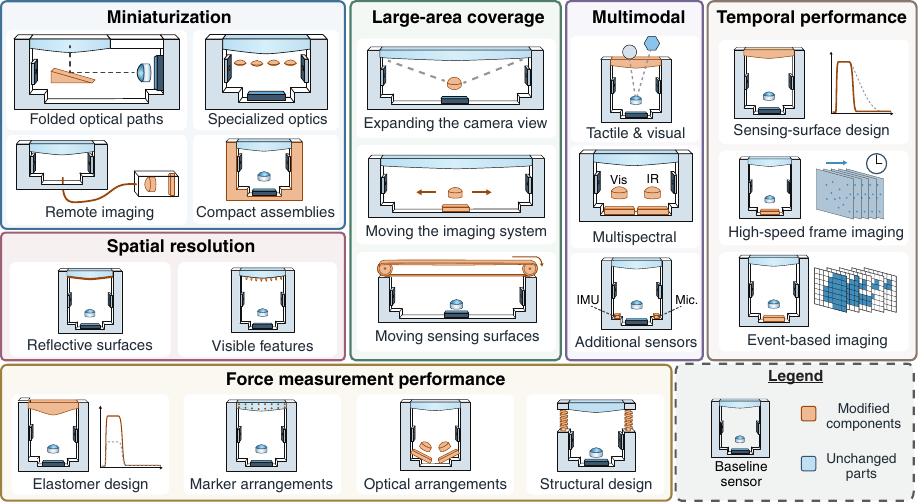}
\caption{Illustrative hardware design approaches organized by six goals. Two-dimensional schematics show physical changes relative to the baseline sensor in Fig.~\ref{fig:vbts_mechanics_pipeline}(a). Muted brown highlights modified or added components; unchanged components are shown in gray and blue. Multiple approaches can be combined within one sensor.}
\label{fig:hardware_design_approaches}
\end{figure*}

\begin{figure*}[tp]
\centering
\includegraphics[width=\textwidth,height=0.88\textheight,keepaspectratio]{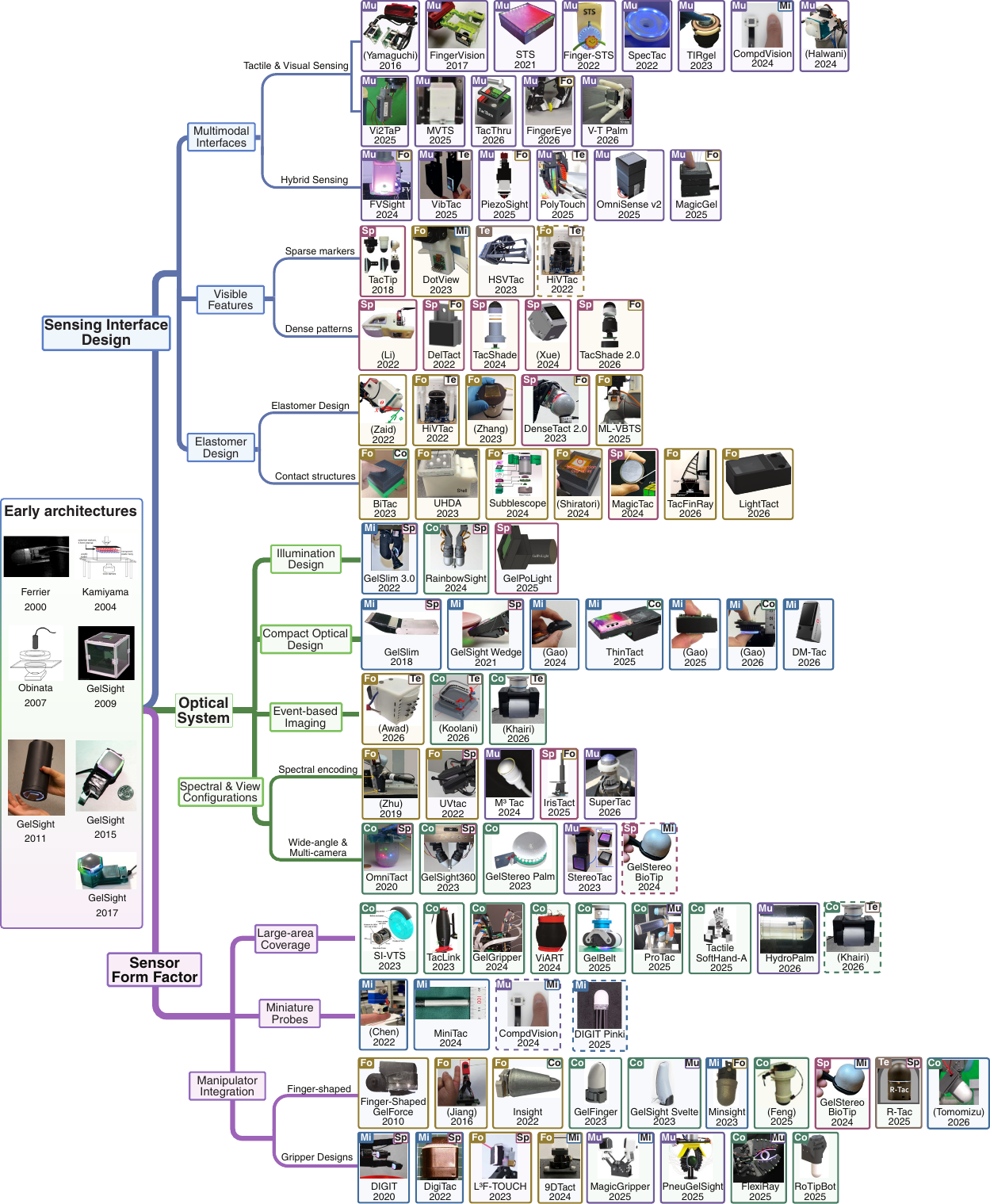}
\caption{Component-based taxonomy of representative VBTS hardware organized by early architectures, sensing-interface design, optical system, and sensor form factor. Colored badges annotate the primary design goals addressed by each implementation: miniaturization (Mi), large-area coverage (Co), spatial resolution (Sp), force measurement performance (Fo), temporal performance (Te), and multimodal sensing (Mu). Dashed borders denote sensors that span multiple categories and appear under more than one branch.}
\label{fig:hardware_taxonomy_overview}
\end{figure*}

From existing hardware designs, we identify six design goals: 1) \textbf{miniaturization}: reducing sensor dimensions; 2) \textbf{large-area coverage}: sensing contact over a larger area; 3) \textbf{spatial resolution}: capturing fine contact details; 4) \textbf{force measurement performance}: estimating forces accurately across the intended range; 5) \textbf{temporal performance}: capturing rapid contact changes; and 6) \textbf{multimodal sensing}: acquiring complementary types of information. For each goal, we summarize design approaches reported in the literature and examine their hardware choices and trade-offs, including those between sensing performance and calibration effort. To connect these goals with concrete devices, Figs.~\ref{fig:hardware_design_approaches} and~\ref{fig:hardware_taxonomy_overview} provide two complementary views. Fig.~\ref{fig:hardware_design_approaches} takes a goal-driven view, using schematics to show how baseline VBTS components are modified or added for each goal. In parallel, Fig.~\ref{fig:hardware_taxonomy_overview} presents a component-based taxonomy of VBTS implementations across sensing interface, optical system, and form factor, annotating each with its corresponding design goals. This taxonomy also supports subsystem-level modification, allowing designers to identify relevant sensors and design options for a target component.

Importantly, hardware design and tactile data interpretation form a closed feedback loop: design goals guide physical choices that shape the observations available for calibration and estimation, while evaluating measurement and task errors exposes physical bottlenecks that motivate hardware or calibration redesign.

\subsection{Miniaturization}
\label{subsec:codesign_compactness}
Miniaturization reduces sensor dimensions so that a VBTS can fit inside anthropomorphic fingertips, slender parallel-jaw fingers, or confined inspection spaces. The principal physical bottleneck is optical: a conventional camera requires a minimum working distance between the lens and the elastomer to focus and cover the sensing area, while light sources must illuminate that surface without creating harsh glare or dark zones. Existing designs overcome this geometric constraint through folded optical paths, specialized optical elements, remote imaging, and compact sensor assemblies, which primarily modify the imaging optics, image sensor, and sensor architecture.

\textbf{Folded optical paths.} Mirrors can redirect the optical path along the finger axis rather than perpendicular to the contact pad, allowing the required working distance to fit within a slim finger profile. GelSlim and GelSight Wedge use folded paths in tactile fingers~\cite{gelslimhighresolutioncompactrobustcalibratedtactilesensingfinger2018,gelsightwedgemeasuringhighresolution3dcontactgeometrycompactrobotfinger2021}. This approach changes the imaging optics and sensor architecture while requiring illumination to fit within the narrow cavity and reach the sensing surface evenly; mirror alignment tolerances and foreshortening across the tilted view also increase geometric calibration effort.

\textbf{Specialized optical elements.} Alternative optical elements replace a single long-throw lens with flatter imaging structures to shorten the axial thickness. A microlens array provides multiple local views in a thin sensor~\cite{thinformatvisionbasedtactilesensormicrolensarraymla2022}. ThinTact instead uses a coded mask for lensless imaging; the mask was selected using simulated image-reconstruction quality and tested in the fabricated sensor, while its optical encoding requires mask-specific calibration and computational image reconstruction~\cite{thintactthinvisionbasedtactilesensorlenslessimaging2025}. Such elements sharply reduce thickness, but shift complexity from mechanical space into optical calibration and image-decoding computation.

\textbf{Remote imaging.} Fiber bundles allow the camera and associated electronics to be placed away from the sensing tip, removing rigid imaging hardware from the distal contact site. DIGIT Pinki and MISTac relay tactile images through imaging fibers to a remote camera; MISTac integrates an 8-mm replaceable tip with separate illumination fibers~\cite{usingfiberopticbundlesminiaturizevisionbasedtactilesensors2025,koch2026mistac}. A related 1.7-mm optical force--torque probe uses a coherent bundle to relay deformation-dependent light patterns to a camera for force and torque estimation, rather than directly imaging surface shape~\cite{zhang2026deformationencoded}. These designs minimize distal size and electrical wiring at the tip, while finite fiber-core counts, honeycomb sampling artifacts, light attenuation, and minimum bend radii constrain image fidelity and routing along articulated joints.

\textbf{Compact sensor assemblies.} Miniaturization can also be achieved by closely co-packaging miniature cameras, light guides, and elastomers within a tight housing. DIGIT uses a compact housing with a replaceable sensing pad, while MiniTac encloses its sensing surface, illumination, and camera in an 8-mm probe~\cite{digitnoveldesignlowcostcompacthighresolutiontactilesensorapplicationinhandmanipulation2020,minitacultracompact8mmvisionbasedtactilesensorenhancedpalpationrobotassistedminimallyinvasivesurgery2024}. GelSlim 3.0 uses a shaped illumination element to distribute LED light across the sensing surface within a thin finger~\cite{gelslim30highresolutionmeasurementshapeforceslipcompacttactilesensingfinger2022}. Minsight miniaturizes the earlier Insight design while retaining all-around sensing by fitting the camera, illumination, and supporting skeleton within a compact fingertip and adapting the elastomer for the shorter camera--surface distance~\cite{minsightfingertipsizedvisionbasedtactilesensorroboticmanipulation2023}. Such designs coordinate the housing with the elastomer and optics, but shorter camera--surface distances intensify lens distortion, restrict the usable field of view, and make uniform directional illumination harder to achieve.

\subsection{Large-area Coverage}
\label{subsec:codesign_coverage}
Large-area coverage aims to obtain tactile information across an extended surface, such as a whole finger, palm, manipulator link, or large workpiece. A key architectural distinction separates \emph{instantaneous coverage}---observing an entire extended surface in parallel within each frame---from \emph{sequential coverage}, which scans a localized optical view across a larger region over time. Existing designs pursue these options by expanding the camera view, moving the imaging system to acquire successive views, or moving the sensing surface across an object.

\textbf{Expanding the camera view.} A wider camera view or multiple views can make more of an extended sensing surface instantaneously observable. GelSight Svelte uses mirrors to direct a single camera's view along a finger-shaped sensing surface, while GelSight360 uses wide-angle optics to observe a curved surface~\cite{gelsightsveltehumanfingershapedsinglecameratactilerobotfingerlargesensingcoverageproprioceptivesensing2023,gelsight360omnidirectionalcamerabasedtactilesensordexterousroboticmanipulation2023}. ViART uses a single fisheye camera to observe markers around the equator of a cylindrical sensing surface~\cite{viartvisionbasedsofttactilesensingautonomousroboticvehicles2024}. Other designs use multiple cameras: OmniTact distributes five cameras around a fingertip, while ProTac places cameras at both ends of an elongated cylindrical surface~\cite{omnitactmultidirectionalhighresolutiontouchsensor2020,visionbasedproximitytactilesensingrobotarmsdesignperceptioncontrol2025}. BiTac uses stereo cameras to observe a 100-mm-square marked membrane, whereas an event-based tactile skin uses two cameras to localize contact on a large optical waveguide~\cite{bitacsoftvisionbasedtactilesensorbidirectionalforceperceptionrobots2023,eventbasedoptotactileskin2026}. Expanding coverage must be coordinated with surface curvature, resolution budget, and illumination: spreading fixed camera pixels over a larger curved shell lowers spatial resolution and introduces severe perspective foreshortening near grazing angles, while multi-camera designs increase wiring, synchronization, and calibration overhead. For example, GelSight360's internal light boards occlude part of its camera view, whereas RainbowSight places LEDs around the base to illuminate curved surfaces without those boards~\cite{gelsight360omnidirectionalcamerabasedtactilesensordexterousroboticmanipulation2023,rainbowsightfamilygeneralizablecurvedcamerabasedtactilesensorsshapereconstruction2024}. The resulting observations can also require position-dependent calibration. GelStereo Palm and BioTip, for instance, calibrate stereo cameras and account for refraction through their curved interfaces~\cite{gelstereopalmnovelcurvedvisuotactilesensor3dgeometrysensing2023,gelstereobiotipselfcalibratingbionicfingertipvisuotactilesensorroboticmanipulation2024}.

\textbf{Moving the imaging system.} A camera or optical module can move beneath a fixed sensing surface to observe different regions in succession without sacrificing local magnification. Combining these views requires their images to be aligned in a common coordinate system. GelFinger rotates its camera and compares two ways of aligning the views: one uses pre-calibrated image transformations, while the other adjusts the alignment locally to reduce errors caused by changes in viewpoint, at greater computational cost~\cite{gelfingernovelvisualtactilesensormultiangletactileimagestitching2023}. Gao et al. track markers at the edge of a stationary elastomer to estimate the rotation center and the angle of each frame, then align the images into a common view; separate spherical indentations calibrate surface reconstruction~\cite{thinpinholebasedtactilesensorrotationalscanningfieldofviewexpansion2026}. These methods preserve high local image quality across a larger area, but mechanical actuation, image stitching, and sequential acquisition introduce latency and limit detection of simultaneous, fast contacts outside the current field of view.

\textbf{Moving sensing surfaces.} A belt or roller can inspect a larger region through continuous rolling contact without instrumenting the entire area at once. GelBelt combines a rolling silicone belt and supporting rollers with fixed optics and illumination~\cite{gelbeltvisionbasedtactilesensorcontinuoussensinglargesurfaces2025}. An event-based tactile roller similarly moves its sensing surface over an object while recording contact changes~\cite{theyseemerollinghighspeedeventvisionbasedtactilerollersensorlargesurfaceinspection2026}. These designs extend the area inspected over time rather than the area observed in a single image. Reconstructing a global surface map requires registering successive frames or events accurately, and belt slip, scanning speed, and contact preload directly affect both spatial fidelity and registration drift.

\subsection{Spatial Resolution}
\label{subsec:codesign_spatial}
Spatial resolution describes the finest geometric or textural detail that a sensor can distinguish, whereas contact localization concerns the accuracy with which it estimates contact position. In a VBTS, effective spatial resolution is not determined by camera pixel count alone: the thickness and stiffness of the deformable elastomer mechanically low-pass filter sharp surface features, the optical system imposes diffraction and defocus limits, and the choice of surface encoding governs what local cues reach the image sensor. Consequently, sensors optimized for fine texture reconstruction and sensors optimized for precise contact localization often adopt distinct surface designs combining coatings, visible features, illumination, and supporting structures.

\textbf{Reflective surfaces.} A thin reflective coating paired with multi-directional lighting converts fine local surface-slope variations into pixel-level shading changes while isolating the camera from the contacted object's color and ambient lighting~\cite{retrographicsensingmeasurementsurfacetextureshape2009,gelsighthighresolutionrobottactilesensorsestimatinggeometryforce2017}. Coating thickness and specular properties directly shape resolution: overly thick or rigid skins smooth out micro-textures, whereas overly specular coatings cause directional saturation. The improved GelSight fingertip combines a matte coating with more uniform illumination to reduce ambiguity in surface-normal estimation~\cite{improvedgelsighttactilesensormeasuringgeometryslip2017}. GelPoLight uses point lights to relax constraints on illumination placement~\cite{gelpolightnovelvisualtactilesensorbasedphotometricstereopointlighting2025}. Agarwal et al. simulated different illumination designs and assessed how well the resulting tactile images could support surface-shape reconstruction, then incorporated a selected design into a physical sensor~\cite{visionbasedtactilesensordesignusingphysicallybasedrendering2025}. Because shading depends nonlinearly on viewing angle, light geometry, and coating reflectance, these designs require calibration relating observed brightness to surface orientation~\cite{3dcalopensourcesoftwarelibrarydepthreconstructionvisionbasedtactilesensors2026}. For example, GelPoLight's spatially varying illumination requires calibration of light positions, intensity, and surface reflectance~\cite{gelpolightnovelvisualtactilesensorbasedphotometricstereopointlighting2025}.

\textbf{Visible features.} Pins, discrete markers, and printed patterns encode surface deformation as trackable spatial primitives rather than continuous shading. Although discrete features sample the contact patch more coarsely than per-pixel photometric stereo, pooling the coupled motion of multiple markers or pins across a compliant skin can yield sub-millimeter contact localization accuracy exceeding the physical spacing between adjacent features---a form of tactile hyperacuity. TacTip uses internal pins to track surface deformation and estimate the location of contact~\cite{tactipfamilysoftopticaltactilesensors3dprintedbiomimeticmorphologies2018,superresolutionopticaltactilesensor2015,additionbiomimeticfingerprintartificialfingertipenhancestactilespatialacuity2017}. Other sensors use dense color patterns, coded markers, or multilayer grids to increase feature density and disambiguate correspondence under large shear~\cite{deltactvisionbasedtactilesensorusingdensecolorpattern2022,3ddensereconstructionvisionbasedtactilesensorcodedmarkers2024,magictacnovelhighresolution3dmultilayergridbasedtactilesensor2024}. DelTact compared tracking errors across printed patterns and selected a pattern that balanced feature density with printing quality, showing how tactile-image measurements can guide feature design~\cite{deltactvisionbasedtactilesensorusingdensecolorpattern2022}. Increasing feature density, however, introduces a direct trade-off: closely spaced markers become harder to track unambiguously under large deformation and can obscure fine surface shading or external views. To address this trade-off, UVtac uses fluorescent markers that are visible under ultraviolet illumination but transparent under white light, enabling marker tracking and a clearer view of the contact outline in alternating modes~\cite{uvtacswitchableuvmarkerbasedtactilesensingfingereffectiveforceestimationobjectlocalization2022}. Ou et al. instead use learning-based methods to estimate markerless images from images with markers, or marker motion from markerless images~\cite{markermarkerlessmodeswitchableopticaltactilesensingdiverserobottasks2024}.

\subsection{Force Measurement Performance}
\label{subsec:codesign_force}
Force measurement performance depends on how sensitively, linearly, and accurately a sensor responds to normal, shear, and torsional loads across its intended operating range. Two physical trade-offs govern this goal. First, \emph{sensitivity competes with measurement range}: a highly compliant elastomer produces large, easily observed deformation under light contact, but bottoms out or enters severe nonlinear stiffening under heavy loads. Second, \emph{distributed force-field resolution competes with decoupled net-load accuracy}: local surface cues resolve spatial pressure distributions but couple strongly to unknown object geometry, whereas compliant structural mounts cleanly isolate net six-axis forces and torques without resolving fine spatial stress fields. Existing designs navigate these trade-offs by modifying the elastomer, marker arrangement, imaging system, or supporting structure.

\textbf{Elastomer design.} Material properties and geometry determine how contact forces deform the sensing surface. ML-VBTS combines a soft outer layer with a harder inner layer to balance sensitivity to small forces with a wider measurement range; its layer properties were compared using simulation and experiments~\cite{enhancingsensitivitymeasurementrangemultilayeredvisionbasedtactilesensormlvbtsparametricstudycomparativebenchmarking2025}. Insight instead uses a locally thinner elastomer region to increase sensitivity to light contact~\cite{softthumbsizedvisionbasedsensoraccurateallroundforceperception2022}. Other designs introduce engineered local structures that amplify deformation or optical contrast under small loads: Subblescope embeds an air bubble in a thin film, while a micro-suction-cup sensor estimates normal force from the deformation of individual cups~\cite{subblescopenovelthinfilmhapticsensingusingsinglebubbleapproach2024,developmentvisionbasedtactilesensormicrosuctioncups2024}. Micro-VBTS uses laser-cut trenches that open under load, producing visible changes in transmitted light for small normal forces~\cite{shi2026microvbts}. Because engineered microstructures introduce nonlinear and contact-dependent stiffness, their optical responses must be calibrated across realistic loading conditions. For example, Subblescope compares changes in the bubble contour with forces measured by a testing machine and calibrates shear response both at initial contact and under a normal preload~\cite{subblescopenovelthinfilmhapticsensingusingsinglebubbleapproach2024}.

\textbf{Marker arrangements.} The number, positions, and depths of markers determine which three-dimensional deformation modes are observable. Placing markers at multiple depths separates normal compression from tangential shear, while sparse layouts simplify tracking at the cost of coarser spatial pressure resolution. Kamiyama et al. compare simulated force-reconstruction errors to select the separation of two marker layers; Finger-Shaped GelForce adapts a two-layer arrangement to a curved fingertip~\cite{evaluationvisionbasedtactilesensor2004a,fingershapedgelforcesensormeasuringsurfacetractionfieldsrobotichand2010}. Sferrazza and D'Andrea instead embed a random marker pattern in the elastomer to support normal-force distribution estimation~\cite{sferrazza2019fullresolution}. HiVTac positions four markers to balance sensitivity to normal and shear forces within a small contact region~\cite{hivtachighspeedvisionbasedtactilesensorpreciserealtimeforcereconstructionfewermarkers2022}. Tang and Nanayakkara compare marker shapes and arrangements for normal-force estimation and use the observed deformation patterns to guide a concentric-ring design~\cite{tang2026markerinformation}. These examples show how force-estimation results can inform marker design. Calibrating marker-based force estimation requires measurements of marker motion under known forces; for example, Finger-Shaped GelForce relates marker movements to reference forces applied at different positions and directions~\cite{fingershapedgelforcesensormeasuringsurfacetractionfieldsrobotichand2010}.

\textbf{Camera and optical arrangements.} Additional viewpoints directly resolve out-of-plane indentation and three-dimensional marker motion that are ambiguous from a single monocular projection. BiTac uses stereo cameras to observe a marked membrane and estimate bidirectional normal forces~\cite{bitacsoftvisionbasedtactilesensorbidirectionalforceperceptionrobots2023}. Tac3D uses mirrors and one camera to create two virtual views of markers for three-dimensional deformation and force-distribution estimation~\cite{zhang2022tac3d}. CompdVision uses a microlens array to obtain multiple close-up tactile views for three-axis force estimation~\cite{compdvisioncombiningnearfield3dvisualtactilesensingusingcompactcompoundeyeimagingsystem2024}. These configurations improve geometric observability but require multi-view calibration to recover three-dimensional deformation from images before mapping it to force. For example, BiTac uses known marker positions to calibrate the stereo geometry and lens distortion, then fits a membrane-model parameter using measured forces and displacements~\cite{bitacsoftvisionbasedtactilesensorbidirectionalforceperceptionrobots2023}.

\textbf{Structural design.} Sensor structures support force estimation by channeling contact loads through known compliant elements whose rigid-body displacement can be measured optically, decoupling net force estimation from local contact shape. L$^3$F-TOUCH suspends its gel and camera on springs and observes their motion relative to the sensor base to estimate three-axis force, while Visiflex uses a spring-supported fingertip to measure force and torque from its motion~\cite{l3ftouchwirelessgelsightdecoupledtactilethreeaxisforcesensing2023,fernandez2021visiflex}. Estimating force from this motion requires calibration against known loads; for example, L$^3$F-TOUCH pairs reference forces with the observed suspension motion to calibrate three-axis force estimation, while separate sphere indentations calibrate surface-shape reconstruction~\cite{l3ftouchwirelessgelsightdecoupledtactilethreeaxisforcesensing2023}. FingerEye instead supports a rigid transparent contact plate with a compliant ring and calibrates the plate's motion against forces and torques~\cite{fingereyecontinuousunifiedvisiontactilesensingdexterousmanipulation}. Minsight compares force and contact-location errors for different internal skeletons and selects a sparse structure for its intended force range; a separate study uses simulated force-reconstruction errors to select elastomer thickness and boundary constraints~\cite{minsightfingertipsizedvisionbasedtactilesensorroboticmanipulation2023,improvingforcereconstructionperformancevisionbasedtactilesensorsoptimizingelasticbody2023}. LightTact combines a soft transparent contact surface with a rigid viewing window to detect gentle contact optically without relying on macroscopic deformation~\cite{lighttactvisualtactilefingertipsensordeformationindependentcontactsensing2026}.

\subsection{Temporal Performance}
\label{subsec:codesign_temporal}
Temporal performance describes how quickly a sensor responds to rapid contact transients---such as impact, vibration, texture scanning, and incipient slip---and delivers actionable estimates to a controller. End-to-end temporal response is governed by three cascaded stages: (i) the \emph{mechanical bandwidth} of the elastomer, whose viscoelastic relaxation and recovery time place a physical ceiling on repeatable dynamic sensing regardless of camera speed; (ii) the \emph{optical acquisition rate and data bandwidth} of the image sensor; and (iii) the \emph{computational latency} of marker tracking, photometric integration, or neural inference. Existing designs target these bottlenecks through sensing-surface design, high-speed frame imaging, and event-based imaging.

\textbf{Sensing-surface design.} The sensing surface must deform promptly on contact and rebound quickly upon release without lingering viscoelastic drift. GelPoLight selects its silicone mixture to balance sensitivity with recovery time~\cite{gelpolightnovelvisualtactilesensorbasedphotometricstereopointlighting2025}. PolyTouch offers interchangeable sensing surfaces and notes that its silicone surface recovers faster than its viscoelastic acrylic-foam alternative~\cite{polytouchrobustmultimodaltactilesensorcontactrichmanipulationusingtactilediffusionpolicies2025}. TacShade 2.0 exhibited delayed surface recovery and errors during sliding; a harder elastomer was proposed to improve recovery~\cite{tacshade20newgreyscalepatternbasedopticaltactilesensorroboticperceptionadaptiveadmittancecontrol2026}. These examples highlight a fundamental trade-off between soft compliance for gentle-contact sensitivity and elastic stiffness for rapid mechanical recovery. Surface encoding also affects computational latency: HiVTac places four markers on a thin elastomer to minimize image-processing cost while calibrating marker displacements against reference force magnitudes and directions~\cite{hivtachighspeedvisionbasedtactilesensorpreciserealtimeforcereconstructionfewermarkers2022}.

\textbf{High-speed frame imaging.} Faster cameras and compact image representations shorten acquisition latency and reduce motion blur during high-frequency vibration or sliding, at the cost of higher illumination intensity requirements (to maintain signal-to-noise ratio under short exposures) and greater data-transfer bandwidth. HSVTac uses a high-speed camera to observe marker motion during vibration and sliding~\cite{hsvtachighspeedvisionbasedtactilesensorexploringfingertiptactilesensitivity2023}. A modified TacTip uses a 120-fps camera to track internal pins for slip detection, while Kato et al. use a 601-fps camera to track markers for force estimation during robotic palpation~\cite{james2018biomimeticslip,kato2024highspeedpalpation}. R-Tac combines a monochrome camera with monochrome illumination to reduce image data volume for shape sensing and calibrates image intensity against known surface deformation~\cite{rtacroundedmonochromevisionbasedtactilesensor2025}.

\textbf{Event-based imaging.} Event cameras report per-pixel brightness changes asynchronously with microsecond-scale temporal resolution rather than capturing redundant full frames at fixed intervals. NeuroTac observes the internal pins of a soft fingertip with an event camera, while Evetac uses an event camera to track markers and detect rapid vibration and slip~\cite{neurotacneuromorphicopticaltactilesensorappliedtexturerecognition2020,evetaceventbasedopticaltactilesensorroboticmanipulation2024}. Other designs use event cameras to observe contact on an optical skin, estimate force and slip, or scan a surface with a roller~\cite{eventbasedoptotactileskin2026,novelmodelbasedforceobjectslipestimationapproachneuromorphicvisiontactilesensors2026,theyseemerollinghighspeedeventvisionbasedtactilerollersensorlargesurfaceinspection2026}. Event sensing drastically reduces data redundancy during static contact and eliminates conventional frame motion blur, but introduces distinct challenges: sustained static loads generate few or no events once deformation stabilizes, sensor output is sensitive to contrast-threshold variation and noise, and asynchronous event streams require specialized temporal processing or reconstruction algorithms. Event-camera calibration can characterize the brightness change required to trigger an event; for example, a general calibration method estimates each pixel's threshold and bias by comparing its events with simultaneously recorded intensity images~\cite{wang2020eventthresholdcalibration}.

\subsection{Multimodal Sensing}
\label{subsec:hardware_multimodal}
Multimodal designs combine tactile images with complementary physical signals---such as pre-contact visual appearance, proximity, temperature, high-frequency vibrotactile acoustics, or auxiliary force cues---within one sensing module. Integrating multiple modalities within a compact fingertip introduces cross-modal interference, spatial-registration demands, and multi-rate synchronization requirements. Existing designs address these integration challenges by combining tactile and visual sensing, exploiting multiple spectral channels, or embedding non-optical transducers.

\textbf{Integrated tactile and visual sensing.} Transparent contact surfaces allow a single camera or dual-view optical module to observe pre-contact scene geometry and object appearance before touch, then transition smoothly to contact-induced deformation upon touch. FingerVision uses a transparent marked elastomer, whereas See-Through-Your-Skin combines a clear gel with a semitransparent reflective layer~\cite{implementingtactilebehaviorsusingfingervision2017,seeingyourskinrecognizingobjectsnovelvisuotactilesensor2021}. TacThru uses ring markers and continuous illumination to track deformation while retaining an external view, while FingerEye combines a transparent rigid cover, a compliant ring, and two camera views~\cite{simultaneoustactilevisualperceptionlearningmultimodalrobotmanipulation2026,fingereyecontinuousunifiedvisiontactilesensingdexterousmanipulation}. Other designs actively switch between tactile and visual observations: SpecTac switches ultraviolet LEDs to reveal fluorescent markers for tactile sensing, whereas Vi2TaP rotates polarizing films to select between tactile and proximity sensing~\cite{spectacvisualtactiledualmodalitysensorusinguvillumination2022,vi2tapcrosspolarizationbasedmechanismperceptiontransitiontactileproximitysensingapplicationssoftgrippers2025}. Continuous see-through designs expose tactile feature extraction to external background clutter and ambient lighting changes, while dense markers partially occlude the external view; conversely, mode-switching mechanisms isolate the two channels at the cost of switching latency and reduced effective frame rate. The tactile and visual outputs also require separate optical and mechanical calibration procedures.

\textbf{Multispectral imaging.} Assigning distinct sensing modalities to non-overlapping wavelength bands reduces cross-channel optical interference through coordinated selection of emitters, filters, coatings, and image sensors. M$^3$Tac combines visible, near-infrared, and thermal imaging to observe external appearance, contact deformation, and temperature~\cite{m3tacmultispectralmultimodalvisuotactilesensorbeyondhumansensorycapabilities2024}. SuperTac extends this approach with ultraviolet-excited fluorescent markers alongside visible, near-infrared, and thermal imaging; it also incorporates triboelectric and inertial sensing~\cite{biomimeticmultimodaltactilesensingenableshumanlikeroboticperception2026}. OmniSense v2 uses programmable ultraviolet and visible illumination with fluorescent markers, while thermochromic patches encode temperature through visible color changes~\cite{omnisensev2humanskininspiredvisuotactilesensorunifiedtactileimaging2025}. Separating modalities spectrally preserves simultaneous acquisition, but adds optical complexity, packaging bulk, and channel-specific calibration: for example, M$^3$Tac aligns its camera views and calibrates temperature, whereas OmniSense v2 uses reference colors to classify the temperature-sensitive patches~\cite{m3tacmultispectralmultimodalvisuotactilesensorbeyondhumansensorycapabilities2024,omnisensev2humanskininspiredvisuotactilesensorunifiedtactileimaging2025}.

\textbf{Additional sensing components.} Non-optical transducers complement camera-based deformation sensing in regimes where optical imaging is bandwidth-limited or mechanically insensitive, such as kHz-range micro-vibrations during texture exploration, acoustic impact transients, or direct electrical contact pressure. VibTac combines a microphone and an accelerometer with optical imaging, while PolyTouch integrates a contact microphone into its tactile finger~\cite{vibtachighresolutionhighbandwidthtactilesensingfingermultimodalperceptionroboticmanipulation,polytouchrobustmultimodaltactilesensorcontactrichmanipulationusingtactilediffusionpolicies2025}. A microphone has also been embedded in a TacTip to capture vibrations during texture exploration~\cite{pestell2022artificialvibrotactile}. PiezoSight instead combines a piezoresistive layer with optical markers~\cite{piezosightcouplingvisionbasedtactilesensorneuralnetworkprocessingpiezoresistivestimulationenhancedpiezovisionhybridsensing2025}. Incorporating heterogeneous transducers requires careful mechanical placement so added components do not obstruct the optical path or stiffen the elastomer unevenly, together with separate physical calibration and cross-rate timestamp synchronization; for example, VibTac calibrates its accelerometer separately from its optical shape measurement~\cite{vibtachighresolutionhighbandwidthtactilesensingfingermultimodalperceptionroboticmanipulation}.

Across these six design goals, hardware choices rarely act independently and instead form a coupled design space summarized in Table~\ref{tab:hardware_tradeoffs}. Miniaturizing optical paths constrains field of view and illumination uniformity; expanding coverage across curved surfaces reduces per-area pixel density and increases perspective distortion; softer elastomers improve light-force sensitivity at the expense of measurement range and temporal recovery; and dense markers, opaque reflective coatings, or auxiliary transducers can occlude fine texture or external visual views. Moreover, nearly every hardware modification shifts the burden of calibration and computation---from mirror and refraction geometry to mask-specific decoding, viscoelastic compensation, or cross-modal synchronization. Navigating these coupled physical trade-offs motivates the calibration, representation learning, simulation, and multimodal control strategies examined in Section~\ref{sec:data_learning_regimes}.

\begin{table*}[t]
\centering
\caption{Summary of the six VBTS hardware design goals in Section~\ref{sec:hardware_taxonomy}, linking primary physical levers to their main calibration and algorithmic requirements and coupled engineering trade-offs.}
\label{tab:hardware_tradeoffs}
\small
\renewcommand{\arraystretch}{1.14}
\setlength{\tabcolsep}{4pt}
\begin{tabularx}{\textwidth}{@{}P{0.15\textwidth}P{0.25\textwidth}P{0.28\textwidth}Y@{}}
\toprule
\textbf{Design goal} & \textbf{Primary physical levers} & \textbf{Calibration and computational impact} & \textbf{Key coupled trade-offs} \\
\midrule
Miniaturization (Sec.~\ref{subsec:codesign_compactness}) & Folded mirrors, microlens arrays, coded masks, fiber bundles, compact housings & Mirror alignment, lens distortion, mask-specific image decoding, fiber-bundle spatial mapping & Reduced thickness vs.\ field of view, illumination uniformity, and optical image fidelity \\
Large-area coverage (Sec.~\ref{subsec:codesign_coverage}) & Wide-angle/fisheye optics, multi-camera arrays, rotating cameras, belts/rollers & Multi-view and curved-refraction calibration; frame stitching and rolling registration & Larger sensing area vs.\ local spatial resolution, foreshortening, bulk, and sequential scan delay \\
Spatial resolution (Sec.~\ref{subsec:codesign_spatial}) & Thin matte reflective coatings, directional lighting, pins, dense/coded or UV markers & Photometric reflectance calibration; marker tracking and correspondence under large shear & Fine texture/shading detail vs.\ marker occlusion, coating durability, and external transparency \\
Force measurement (Sec.~\ref{subsec:codesign_force}) & Multilayer or microstructured elastomers, multi-depth markers, stereo views, compliant mounts & Nonlinear load--displacement fitting, stereo triangulation, decoupled suspension calibration & Light-force sensitivity vs.\ load range and hysteresis; net 6-axis accuracy vs.\ spatial force fields \\
Temporal performance (Sec.~\ref{subsec:codesign_temporal}) & Fast-recovering elastomers, sparse markers, high-speed/monochrome or event cameras & Contrast-threshold characterization, asynchronous event tracking, short-exposure lighting & High dynamic bandwidth vs.\ elastomer compliance, data rate, and static-load observability in events \\
Multimodal sensing (Sec.~\ref{subsec:hardware_multimodal}) & Transparent skins, switchable UV/polarizers, multispectral channels, acoustic/IMU/piezo elements & Multi-channel spatial alignment, separate physical calibrations, cross-rate synchronization & Complementary pre-contact/vibrotactile/thermal cues vs.\ background clutter, bulk, and mode switching \\
\bottomrule
\end{tabularx}
\end{table*}

\section{Data and Learning Regimes: From Sensor Readout to Embodied Intelligence}
\label{sec:data_learning_regimes}

\begin{figure*}[t]
\centering
\includegraphics[width=\textwidth]{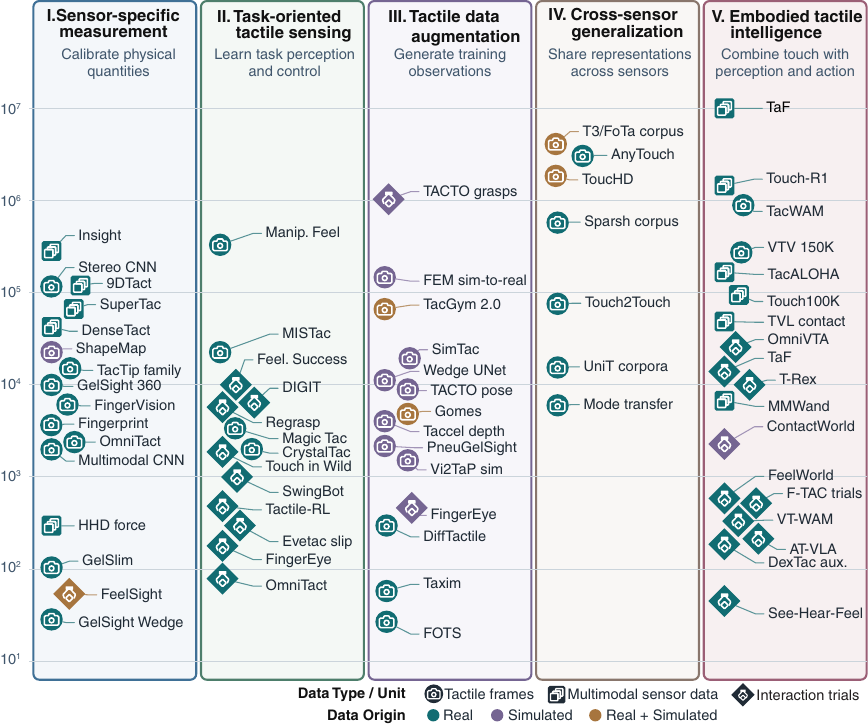}
\caption{Reported data counts across the five data and learning regimes. A shared logarithmic count axis displays optical tactile observation records---distinguishing tactile frames (single-camera icons) and multimodal sensor data (stacked-frame icons)---and robot interaction trials or episodes (diamond icons). An observation record corresponds to one sensor at one retained timestamp; an interaction trial or episode may contain many records, so equal heights do not imply equivalent data volumes. Color distinguishes real (teal), simulated (purple), and real-plus-simulated (brown) data.}
\label{fig:five_stage_roadmap}
\end{figure*}

VBTS hardware provides an interface for sensing physical interactions. The resulting tactile data can support sensing and robotic applications with different data requirements. We organize these uses into five data and learning regimes. Fig.~\ref{fig:five_stage_roadmap} compares reported data budgets using explicit counting units, while Table~\ref{tab:stage_comparison} compares their purposes, data, and example applications. The regimes can overlap and do not imply successively larger datasets.

For each regime, we examine how tactile data are obtained and used for sensing and robotic tasks. In the formulations below, $H_i$ denotes a sensor hardware configuration, $\mathcal{T}_j$ a task, $D$ a dataset, and $M$ a model. The subscripts $i$ and $j$ identify the sensor and task. Superscripts $\mathrm{real}$ and $\mathrm{sim}$ distinguish physical and simulated data. We use $S$ for a simulator, $N$ for the number of sensor hardware configurations, and $Z_{\mathrm{shared}}$ for a shared representation. Arrows indicate the relationships between the hardware, data, and models.

To compare reported dataset budgets consistently in Fig.~\ref{fig:five_stage_roadmap}, we compute the total number of optical observation records for compatible optical data as $N_{\mathrm{obs}}=\sum_g N_g m_g$, where $g$ indexes disjoint data groups, $N_g$ is the reported count in group $g$, and $m_g$ is the number of optical sensor records per reported datum, established from the acquisition or input-recording protocol. Simultaneous camera views of one sensor are combined, whereas distinct physical sensors are counted separately; additional labels or representations of the same record do not increase $N_{\mathrm{obs}}$. For example, 39,197 paired observations from two DIGIT fingertips correspond to 78,394 sensor records~\cite{graspindependentindirecttoolforceestimationusingvisionbasedtactilesensors2026}, while 1,253 tactile videos containing 60 frames each correspond to 75,180 records~\cite{spatiotemporaldualattentionnetworkenabledshapeindependenthardnessestimationvisionbasedtactilesensor2025}. TacWAM's four tasks, 300 episodes per task, approximately 500 steps per episode, and two sensors yield approximately 1.2 million records~\cite{tacwamanchorguidedworldactionmodelmechanicsawaretactileprediction2026}. We retain native units when sequence lengths, sensor multiplicity, or tactile coverage are unresolved, and distinguish simulator throughput from the finite datasets generated in a reported experiment.

\begin{table*}[t]
\centering
\caption{Purposes, data, and example applications of the five tactile data and learning regimes.}
\label{tab:stage_comparison}
\small
\renewcommand{\arraystretch}{1.14}
\setlength{\tabcolsep}{4pt}
\begin{tabularx}{\textwidth}{@{}P{0.14\textwidth}YYY@{}}
\toprule
\textbf{Regime} & \textbf{Purpose} & \textbf{Primary data and supervision} & \textbf{Example application} \\
\midrule
I. Sensor-Specific Measurement & Estimate physical contact quantities & Single-sensor tactile observations with known contact measurements & Force estimation: FeelAnyForce~\cite{feelanyforceestimatingcontactforcefeedbacktactilesensationvisionbasedtactilesensors2025} \\
II. Task-Oriented Tactile Sensing & Guide a specific robotic task & Task-specific tactile interaction data with actions or outcomes & Insertion with unfamiliar geometry~\cite{tactilerlinsertiongeneralizationobjectsunknowngeometry2021} \\
III. Tactile Data Augmentation & Generate data to expand training & Scalable simulated or generated observations, optionally combined with real data & Grasp-stability learning: TACTO~\cite{tactofastflexibleopensourcesimulatorhighresolutionvisionbasedtactilesensors2022} \\
IV. Cross-Sensor Generalization & Transfer tactile knowledge across sensors & Tactile observations from multiple sensor units or designs, with optional target-sensor adaptation data & Shared representations: T3~\cite{transferabletactiletransformersrepresentationlearningdiversesensorstasks2024} \\
V. Embodied Tactile Intelligence & Combine touch with other modalities for perception and action & Tactile data aligned with vision, language, robot states, or actions & Bimanual manipulation: Tactile ALOHA~\cite{tactilealohalearningbimanualmanipulationtactilesensing2025} \\
\bottomrule
\end{tabularx}
\end{table*}

\subsection{Regime I: Sensor-Specific Measurement}
\label{subsec:stage_sensor_specific}

The first regime focuses on estimating contact quantities from sensor-specific tactile observations and assessing measurement accuracy. Tactile observations are collected together with known contact measurements to develop a model for the sensor:
\[
H_i\rightarrow D_i\rightarrow M_i.
\]
For example, recording tactile images together with applied forces provides data for training a force estimator~\cite{feelanyforceestimatingcontactforcefeedbacktactilesensationvisionbasedtactilesensors2025}. The resulting measurements can be used in different robotic tasks.

\subsubsection{\textup{\textbf{Force estimation}}}
Force-estimation methods predict either the net contact wrench or its spatial traction distribution from raw tactile images. Calibrated methods relate observed displacement to measured force and torque, while supervised models learn from tactile images or displacement features recorded under known loads~\cite{effectiveestimationcontactforcetorquevisionbasedtactilesensorshelmholtzhodgedecomposition2019,visionbasedtactilesensormechanismestimationcontactpositionforcedistributionusingdeeplearning2021,3dforceidentificationpredictionusingdeeplearningbasedgelsightstructuredsensor2024}. FeelAnyForce combines force regression with depth supervision and studies adaptation to other optical sensors~\cite{feelanyforceestimatingcontactforcefeedbacktactilesensationvisionbasedtactilesensors2025}. Depth--area models estimate normal force from a more specific relationship between contact depth, contact area, and applied load~\cite{depthareadependentnormalforceestimationmodelvisionbasedtactilesensors2025}. A central challenge in force supervision is that a standard six-axis force--torque transducer measures only the resultant net load rather than the local pressure field. For distributed force estimation, inverse finite-element methods recover contact forces from measured deformation, and finite-element simulations can also provide dense training labels for learned estimators~\cite{sferrazza2019ground,sferrazza2022sim,ifem20dense3dcontactforcefieldreconstructionassessmentvisionbasedtactilesensors2025}. However, these synthetic labels inherit the assumed constitutive model, boundary conditions, and friction parameters; consequently, close agreement with a net force--torque transducer does not by itself verify local traction accuracy across sharp edges or multi-point contacts.

Event-based force estimators use temporal models to interpret brightness changes recorded over time~\cite{noveldynamicvisionbasedapproachtactilesensingapplications2020,dynamicvisionbasedforcemeasurementsusingconvolutionalrecurrentneuralnetworks2020,novelmodelbasedforceobjectslipestimationapproachneuromorphicvisiontactilesensors2026}. Unlike frame-based cameras that directly observe absolute deformation against a reference image, event streams encode only incremental intensity changes; maintaining accurate force estimates during sustained holds therefore requires careful temporal integration, noise filtering, or drift-mitigating mechanical models.

Indirect force estimation uses observations from a sensor holding a tool to infer forces acting at the tool tip. Because the mapping depends on lever arms, grasp pose, and how the tool transmits loads through the fingers, grasp-independent tool-force estimation addresses this dependence with learned models~\cite{graspindependentindirecttoolforceestimationusingvisionbasedtactilesensors2026}. Training data must therefore cover the relevant tools and grasp configurations. Similarly, models for sensors embedded in compliant fingers must distinguish local contact deformation from global deflection of the finger structure.

\subsubsection{\textup{\textbf{Geometry, localization, and property estimation}}}
Local geometry estimation uses tactile images or tracked marker displacement to recover surface shape or three-dimensional displacement. GelSight and 3D Cal illustrate calibrated surface reconstruction, with 3D Cal providing an automated calibration and depth-reconstruction workflow~\cite{gelsighthighresolutionrobottactilesensorsestimatinggeometryforce2017,3dcalopensourcesoftwarelibrarydepthreconstructionvisionbasedtactilesensors2026}. Learned methods use coded markers or multiple prediction objectives to reconstruct spatial deformation~\cite{3ddensereconstructionvisionbasedtactilesensorcodedmarkers2024,realtimereconstruction3dtactilemotionfieldmultitasklearning2024}. Depth-from-defocus and other displacement measurement methods provide additional reconstruction approaches~\cite{improveddfdmethodthreedimensionaldisplacementmeasurementvisionbasedtactilesensor2024,highresolutionreliable3ddeformationmeasurementmethodvisionbasedtactilesensor2025}. Asynchronous event graphs offer another representation for contact angle estimation~\cite{tactigraphasynchronousgraphneuralnetworkcontactanglepredictionusingneuromorphicvisionbasedtactilesensing2023}.

Object reconstruction combines information from multiple contacts. Tac2Structure accumulates local surface observations, while TouchSDF and diffusion-based methods use learned shape priors to complete the object~\cite{tac2structureobjectsurfacereconstructiononlymultitimestouch2023,touchsdfdeepsdfapproach3dshapereconstructionusingvisionbasedtactilesensing2024,endtoenddiffusionbased3dobjectreconstructionrobotictactilesensing2026}. The resulting shape can therefore contain both observed contact regions and inferred surfaces. For localization, MidasTouch estimates the position of sliding contact on a known object, whereas NormalFlow tracks relative motion by aligning surface-normal maps~\cite{midastouchmontecarloinferencedistributionsslidingtouch2022,normalflowfastrobustaccuratecontactbasedobject6dofposetrackingvisionbasedtactilesensors2025}. FingerSLAM, ShapeMap, ViHOPE, and NeuralFeels combine visual and tactile information for localization or reconstruction~\cite{fingerslamclosedloopunknownobjectlocalizationreconstructionvisuotactilefeedback2023,shapemap3defficientshapemappingdensetouchvision2021,vihopevisuotactileinhandobject6dposeestimationshapecompletion2023,neuralfeelsneuralfieldsvisuotactileperceptioninhandmanipulation2024}. Their data requirements differ according to whether object geometry is known and whether shape, pose, or both must be estimated.

Tactile observations also support material and object inference. Fabric exploration, visual--tactile matching, and texture recognition use contact observations to identify materials or objects~\cite{activeclothingmaterialperceptionusingtactilesensingdeeplearning2018,connectinglookfeelassociatingvisualtactilepropertiesphysicalmaterials2017,learningidentifyobjectinstancestouchtactilerecognitionmultimodalmatching2019,vitacfeaturesharingvisiontactilesensingclothtexturerecognition2018}. Sparse feature graphs also provide inputs for tactile classification~\cite{tactilegattactilegraphattentionnetworksrobottactileperceptionclassification2024}. Firmness and hardness estimation relate the sensor response to the probing action and object geometry~\cite{cherrytomatofirmnessdetectionpredictionusingvisionbasedtactilesensor2024,wearablethumbdevicefruitfirmnessestimationvisionbasedtactilesensing2025,spatiotemporaldualattentionnetworkenabledshapeindependenthardnessestimationvisionbasedtactilesensor2025}. A classifier may identify texture patterns without measuring a mechanical property. Estimating properties such as hardness therefore requires data that distinguish material response from changes in shape, load, and interaction history.

\subsubsection{\textup{\textbf{Slip and contact-state estimation}}}
Slip and contact-state estimation use temporal tactile observations to detect relative motion or anticipate contact changes. GelSight tracks marker displacement and uses its spatial variation to indicate partial slip~\cite{measurementshearslipgelsighttactilesensor2015}. A force-and-deformation method identifies sticking regions, estimates object motion, and reconstructs local slip~\cite{incipientslipdetectionmethodvisionbasedtactilesensorbaseddistributionforcedeformation2021}. For soft objects, distributed force estimates and deformation gradients support slip assessment while accounting for object deformation~\cite{incipientslipdetectionmethodsoftobjectsvisionbasedtactilesensor2022}. A local-deformation method selects sticking reference points from relative marker motion and iteratively separates sticking and slipping regions~\cite{robustincipientslipdetectionmethodvisionbasedtactilesensorbasedlocaldeformationdegree2023}.

Learned methods estimate contact states from sequences. With FingerVision, a long short-term memory (LSTM) network classifies marker-displacement sequences, while PixelMotionNet predicts future displacement frames for slip anticipation~\cite{learningdetectpredictcontacteventsvisionbasedtactilesensors2019}. FTFNet combines learned frequency bands with temporal features for slip-direction classification and tactile-signal prediction; its experiments on BioTac and piezoresistive sensors provide a related non-optical comparison~\cite{ftfnetfrequencytimefusionnetworkslippredictiondexterousroboticmanipulation2026}. Across both model-based and learned approaches, obtaining clean ground truth for incipient slip remains substantially harder than labeling gross slip, because micro-sliding begins locally within the contact patch before rigid-body object motion is detectable by external trackers. Using these contact-state estimates to select corrective actions is the focus of the next regime.

\subsection{Regime II: Task-Oriented Tactile Sensing}
\label{subsec:stage_task_specific}

The second regime focuses on using tactile observations to support specific robotic tasks. Data collected during interactions with a particular sensor and task are used to develop a task-specific model:
\[
(H_i,\mathcal{T}_j)\rightarrow D_{ij}\rightarrow M_{ij}.
\]
The model may predict task outcomes, estimate task-relevant states, or output robot actions directly. Its inputs can be calibrated physical estimates, raw tactile images, or compact learned features. A key motivation for this regime is bypassing explicit intermediate physical inversion: when a task depends on complex contact patterns (such as grasp stability or edge alignment), training directly on task supervision avoids compounding errors from imperfect force or geometry calibration. Conversely, because the robot's own actions determine which contacts appear in $D_{ij}$, the resulting model $M_{ij}$ is tightly coupled to the specific sensor unit, gripper geometry, and interaction distribution, making transfer across new tasks or hardware substantially harder than in Regime~I.

\subsubsection{\textup{\textbf{Grasping and tactile feedback control}}}
Tactile data support grasping along a spectrum from discrete outcome assessment to continuous closed-loop regulation. At the discrete level, The Feeling of Success predicts grasp outcomes from visual and tactile observations, while More Than a Feeling uses these observations to select regrasp adjustments prior to lifting~\cite{feelingsuccessdoestouchsensinghelppredictgraspoutcomes2017,morefeelinglearninggraspregraspusingvisiontouch2018}. During execution, slip and contact-event estimates trigger immediate corrective grip-force or motion responses~\cite{slipdetectioncombinedtactilevisualinformation2018,learningdetectpredictcontacteventsvisionbasedtactilesensors2019}. For deformable objects, where excessive force damages the object and insufficient force induces slip, exploratory interactions provide contact cues for grasp selection and adaptive force regulation~\cite{learninggeneralizablevisiontactileroboticgraspingstrategydeformableobjectstransformer2025}.

At the planning level, task-specific predictive control learns forward dynamics directly in tactile observation space so that actions can be optimized toward desired contact configurations. Manipulation by Feel uses predicted tactile images to plan rolling and sliding motions toward tactile goals~\cite{manipulationfeeltouchbasedcontroldeeppredictivemodels2019}. SwingBot uses tactile exploration to infer physical properties such as center of mass and friction for dynamic swing-up manipulation~\cite{swingbotlearningphysicalfeaturesinhandtactileexplorationdynamicswingupmanipulation2020}. In both cases, the learned representation is shaped specifically by the actions and objectives of the target task rather than general-purpose physical reconstruction.

\subsubsection{\textup{\textbf{In-hand and precision manipulation}}}
In-hand manipulation requires tracking rapidly shifting fingertip contacts under severe visual occlusion from the robot's own fingers. RotateIt combines vision, proprioception, and contact-location information from optical fingertips for object rotation~\cite{generalinhandobjectrotationvisiontouch2023}. AnyRotate uses tactile contact features and an observation model to transfer rotation policies across hand orientations~\cite{anyrotategravityinvariantinhandobjectrotationsimtorealtouch2024}. Robotic-play pretraining and continuous visual--tactile fingertips provide further approaches to learning useful features and supporting dexterous interaction~\cite{dexteritytouchselfsupervisedpretrainingtactilerepresentationsroboticplay2023,fingereyecontinuousunifiedvisiontactilesensingdexterousmanipulation}. A recurring design lesson in dexterous manipulation is that low-dimensional contact abstractions---such as binary contact states, contact centroids, or local surface normals---often generalize better and transfer more reliably from simulation than raw RGB tactile frames, provided those abstractions can be estimated accurately and at sufficient rate on the physical hand.

Precision insertion relies on millimeter-scale local geometry and contact-wrench cues to resolve jamming and alignment errors that external cameras cannot observe cleanly. GelSight localization of small parts, tactile reinforcement learning for unfamiliar insertion geometry, and tactile-guided exploration illustrate how local contact feedback guides alignment without full prior object models~\cite{localizationmanipulationsmallpartsusinggelsighttactilesensing2014,tactilerlinsertiongeneralizationobjectsunknowngeometry2021,tactileguidedexplorationpositioninghighprecisionroboticpeginholetasks2026}. Other contact-rich tasks require sensing morphologies and representations tailored directly to their interaction mechanics: RoTipBot uses rotatable sensing fingertips for thin-sheet handling, while FingerVision supports contact-aware vegetable cutting~\cite{rotipbotrobotichandlingthinflexibleobjectsusingrotatabletactilesensors2025,combiningfingervisionopticaltactilesensingreducinghandlingerrorscuttingvegetables2016}. Together, these studies show how task mechanics jointly constrain sensor hardware, contact representation, and control law.

\subsubsection{\textup{\textbf{Learning from demonstrations}}}
Demonstration datasets pair multimodal observations with expert actions, enabling contact-sensitive behaviors to be learned without hand-designed reward functions or explicit analytical contact models. Hierarchical multisensory learning, force-matched imitation, and visuotactile bimanual policies use such data to acquire compliant insertion, fastening, and bimanual skills~\cite{seefeelacthierarchicallearningcomplexmanipulationskillsmultisensoryfusion2019,multimodalforcematchedimitationlearningseethroughvisuotactilesensor2025,tactilealohalearningbimanualmanipulationtactilesensing2025}. Collecting high-quality contact demonstrations, however, faces a practical teleoperation bottleneck: human operators rely heavily on natural haptic feedback to modulate contact forces, so teleoperating a robot without haptic cues often yields inconsistent forces or tentative recoveries. Portable hand-held grippers equipped with VBTSs help overcome this bottleneck by allowing demonstrators to feel contact directly and collect rich interaction data beyond fixed laboratory robots~\cite{touchwildlearningfinegrainedmanipulationportablevisuotactilegripper2025}. Because imitation policies can easily overfit to correlated visual cues when demonstrations follow narrow trajectories, controlled ablations comparing policies with and without tactile feedback---especially under visual occlusion or physical perturbation---are essential to verify that touch genuinely drives closed-loop task performance.

\subsection{Regime III: Tactile Data Augmentation}
\label{subsec:stage_data_scaled}

The third regime focuses on generating tactile data to expand training datasets and reduce physical data collection. Simulated observations can be combined with physical sensor data to train a model:
\[
S(H_i)\rightarrow D_i^{\mathrm{sim}},\qquad (D_i^{\mathrm{real}}\cup D_i^{\mathrm{sim}})\rightarrow M.
\]
A simulator $S$ represents the sensor hardware $H_i$ and generates observations for selected contact conditions. Its outputs may include images, deformation, forces, contact features, or interaction trajectories.

\subsubsection{\textup{\textbf{Tactile simulation approaches}}}
Optical simulators generate tactile images from modeled contact. TACTO couples efficient rendering with robot simulation for perception and manipulation~\cite{tactofastflexibleopensourcesimulatorhighresolutionvisionbasedtactilesensors2022}. Taxim uses calibrated optical examples and a marker-motion model~\cite{taximexamplebasedsimulationmodelgelsighttactilesensors2022}. FOTS combines learned optical mapping, shadow generation, and marker-motion approximation for tactile-motor simulation-to-real learning~\cite{fotsfastopticaltactilesimulatorsim2reallearningtactilemotorrobotmanipulationskills2024}. Their outputs provide training observations without recording every contact on a physical sensor. The required optical detail depends on whether the downstream model uses image appearance, geometry, or another contact representation.

Mechanical simulators model deformation and contact dynamics. Tacchi uses a material-point method for elastomer deformation, while DiffTactile combines a finite-element sensor model with object dynamics, contact mechanics, and a learned optical response~\cite{tacchipluggablelowcomputationalcostelastomerdeformationsimulatoropticaltactilesensors2023,difftactilephysicsbaseddifferentiabletactilesimulatorcontactrichroboticmanipulation2024}. DiffTactile's differentiable formulation supports parameter identification and trajectory optimization. Tacchi 2.0 extends deformation simulation to dynamic contact, and SimTac addresses biomorphic sensor structures~\cite{tacchi20lowcomputationalcostcomprehensivedynamiccontactsimulatorvisionbasedtactilesensors2026,simtacphysicsbasedsimulatorvisionbasedtactilesensingbiomorphicstructures2026}.

Robot-learning environments generate contact trajectories for policy training. Tactile Gym 2.0 compares low-cost optical sensors in simulated control tasks, while Taccel supports large parallel workloads on GPUs~\cite{tactilegym20simtorealdeepreinforcementlearningcomparinglowcosthighresolutionrobottouch2022,taccelscalingvisionbasedtactileroboticshighperformancegpusimulation2025}. SimTacLS addresses large-area sensing, including its deformation, image generation, and learning requirements~\cite{simulationlearningapplicationvisionbasedtactilesensinglargescale2023}.

Datasets and generative models provide additional sources of training observations. ObjectFolder and ObjectFolder 2.0 offer virtual multisensory objects for perception and interaction studies~\cite{objectfolderdatasetobjectsimplicitvisualauditorytactilerepresentations2022,objectfolder20multisensoryobjectdatasetsim2realtransfer2022}. ManiFeel provides a benchmark for visuotactile manipulation-policy learning, while contact-conditioned diffusion generates tactile images from specified contact conditions~\cite{manifeelbenchmarkingunderstandingvisuotactilemanipulationpolicylearning2026,visionbasedtactileimagegenerationcontactconditionguideddiffusionmodel2025}.

\begin{table*}[t]
\centering
\caption{Representative simulation approaches, their primary modeling roles, generated outputs, and target validation criteria.}
\label{tab:simulators}
\small
\renewcommand{\arraystretch}{1.12}
\setlength{\tabcolsep}{4pt}
\begin{tabularx}{\textwidth}{@{}P{0.19\textwidth}YYY@{}}
\toprule
\textbf{Platform or family} & \textbf{Principal modeling role} & \textbf{Generated outputs and applications} & \textbf{Validation target} \\
\midrule
TACTO~\cite{tactofastflexibleopensourcesimulatorhighresolutionvisionbasedtactilesensors2022} & Fast optical rendering coupled to a physics engine & Tactile images for perception and control & Appearance and physical transfer for the selected task \\
Taxim~\cite{taximexamplebasedsimulationmodelgelsighttactilesensors2022} & Example-based optical response and marker-motion synthesis & Sensor-specific image and feature generation & Held-out shape, lighting response, and marker displacement \\
Tacchi / Tacchi 2.0~\cite{tacchipluggablelowcomputationalcostelastomerdeformationsimulatoropticaltactilesensors2023,tacchi20lowcomputationalcostcomprehensivedynamiccontactsimulatorvisionbasedtactilesensors2026} & Elastomer deformation; extension to dynamic interaction & Contact surface and changing deformation & Geometry, material response, and contact history \\
DiffTactile~\cite{difftactilephysicsbaseddifferentiabletactilesimulatorcontactrichroboticmanipulation2024} & Differentiable sensor, object, contact, and optical models & System identification and contact-rich optimization & Identified parameters, forces/displacements, and physical outcomes \\
Tactile Gym 2.0~\cite{tactilegym20simtorealdeepreinforcementlearningcomparinglowcosthighresolutionrobottouch2022} & Tactile robot-learning environments & Policies using contact representations & Observation-model accuracy and real-robot control \\
Taccel~\cite{taccelscalingvisionbasedtactileroboticshighperformancegpusimulation2025} & GPU-parallel tactile robotics & Scalable training contacts and rollouts & Full-loop throughput and transfer at matched real-data budgets \\
SimTacLS / SimTac~\cite{simulationlearningapplicationvisionbasedtactilesensinglargescale2023,simtacphysicsbasedsimulatorvisionbasedtactilesensingbiomorphicstructures2026} & Large-area or biomorphic sensor modeling & Morphology-dependent sensing and learning & Spatial error across the body and geometry-specific transfer \\
\bottomrule
\end{tabularx}
\end{table*}

\subsubsection{\textup{\textbf{Simulation-to-real transfer and adaptation}}}
Simulation-to-real transfer bridges two distinct components of the tactile reality gap: an \emph{optical gap}---arising from unmodeled illumination nonuniformity, internal reflections, lens distortion, sensor noise, and coating wear---and a \emph{mechanical gap}---arising from simplified contact dynamics, linear-elastic or penetration approximations, friction anisotropy, and unmodeled viscoelastic hysteresis. Across the simulation platforms above, a clear fidelity--throughput trade-off emerges: fast optical renderers and GPU-parallel environments generate millions of rollouts for reinforcement learning but simplify elastomer continuum mechanics, whereas finite-element and material-point simulators capture richer deformation and shear at substantially higher computational cost. Choosing the right abstraction level therefore depends on the downstream task: policies consuming geometric or structural abstractions (such as depth maps, contact masks, or marker displacements) often cross the reality gap more reliably than models trained directly on raw simulated pixels.

Finite-element-based learning and physics-based optical rendering explicitly model the physical coupling between elastomer deformation and recorded appearance~\cite{learningsensetouchsimulationsimtorealstrategyvisionbasedtactilesensing2020,sferrazza2022sim,simulationvisionbasedtactilesensorsusingphysicsbasedrendering2021}. Rendering studies also examine how material and illumination choices affect the optical response prior to physical fabrication~\cite{visionbasedtactilesensordesignusingphysicallybasedrendering2025}. Data-based approaches instead reduce the optical gap through synthetic GelSight generation, bidirectional CycleGAN translation, or real-to-sim image translation that maps physical images back into the canonical simulated domain~\cite{generationgelsighttactileimagessim2reallearning2021,bidirectionalsimtorealtransfergelsighttactilesensorscyclegan2022,tactilesimtorealpolicytransferrealtosimimagetranslation2022}. Domain randomization varies optical and mechanical parameters during training to cover a wider range of operating conditions~\cite{domainrandomizationtransferringdeepneuralnetworkssimulationrealworld2017}, though its effectiveness depends on whether the randomized distributions bracket the true physical sensor without making the learning problem overly conservative.

SingleS2R and NeuTac investigate transfer with limited real-data adaptation in visual--tactile and neuromorphic settings, respectively~\cite{singles2rsinglesampledrivensimtorealtransfermultisourcevisualtactileinformationunderstandingusingmultiscalevisiontransformers2024,neutaczeroshotsim2realmeasurementneuromorphicvisionbasedtactilesensors2024}. Even when simulation supplies the bulk of training volume, physical data remain essential at specific stages---for optical or mechanical simulator calibration, target-sensor adaptation, and real-world validation---underscoring why simulated and physical budgets must be tracked separately.

Together, optical rendering, physics-based deformation modeling, and GPU-parallel environments reduce physical collection cost, though real-world transfer still depends on simulator fidelity and target-sensor adaptation. Table~\ref{tab:simulators} summarizes the representative simulation platforms in this regime, comparing their principal modeling roles, generated outputs, and validation targets.

\subsection{Regime IV: Cross-Sensor Generalization}
\label{subsec:stage_cross_sensor}

The fourth regime focuses on transferring tactile knowledge across sensors with different hardware configurations. A common approach learns a shared representation from observations collected with multiple sensors:
\[
\{H_i,D_i\}_{i=1}^{N}\rightarrow Z_{\mathrm{shared}}\rightarrow M.
\]
Unlike external RGB cameras that share standardized projective geometry and natural-image statistics, VBTSs exhibit two tightly coupled domain shifts whenever hardware changes: an \emph{optical domain shift} driven by illumination color, camera viewpoint, reflective coating, and marker pattern, and a \emph{mechanical domain shift} driven by elastomer thickness, stiffness, viscoelasticity, and surface curvature. As a result, identical physical contacts can produce markedly different images and deformation fields across devices. Shared models must absorb these hardware-specific variations while preserving the underlying physical contact cues required by the target task.

\subsubsection{\textup{\textbf{Shared tactile representations}}}
Representation learning separates reusable tactile features from task-specific predictions. Sparsh pretrains encoders via self-supervised masking and self-distillation before evaluating them on downstream tasks in TacBench~\cite{sparshselfsupervisedtouchrepresentationsvisionbasedtactilesensing2024}, while UniT targets data-efficient generalization to unseen objects~\cite{unitdataefficienttactilerepresentationgeneralizationunseenobjects2025}.

T3 combines sensor-specific encoders, a shared transformer, and task-dependent decoders trained on FoTa~\cite{transferabletactiletransformersrepresentationlearningdiversesensorstasks2024}; zero-shot transfer was more favorable for pose estimation than object classification, with fine-tuning improving specific sensor--task pairs.

AnyTouch learns unified static--dynamic representations from aligned multi-sensor data, and AnyTouch 2 extends this framework with the ToucHD dataset toward temporal deformation and force-aware perception~\cite{anytouchlearningunifiedstaticdynamicrepresentationmultiplevisuotactilesensors2024,anytouch2generalopticaltactilerepresentationlearningdynamictactileperception2025}. For compact dataset encoding, Tactile Functasets learns neural implicit representations of tactile datasets~\cite{tactilefunctasetsneuralimplicitrepresentationstactiledatasets2025}.

\subsubsection{\textup{\textbf{Tactile signal and measurement transfer}}}
Signal translation converts observations from one tactile sensor domain into those expected by another. Touch2Touch uses this approach to support downstream manipulation methods~\cite{touch2touchcrossmodaltactilegenerationobjectmanipulation2024}. The translated observations must preserve the relevant contact geometry or other target information. Comparing them with actual observations from the target sensor helps distinguish visual similarity from useful measurement transfer.

GenForce transfers force-sensing knowledge through a shared marker representation~\cite{trainingtactilesensorslearnforcesensingeachother2026}. The relationship between marker motion and force still depends on the mechanical responses included in training and adaptation. Related work transfers supervision between sensing modalities: optical tactile measurements support dense shape reconstruction from magnetic sensors or sparse taxels~\cite{supermagvisionbasedtactiledataguidedhighresolutiontactileshapereconstructionmagnetictactilesensors2025,highresolutionreconstructionnonplanartactilepatternslowresolutiontaxelbasedtactilesensors2025}. Point-spread-function-based targets provide another approach to learning dense tactile patterns from sparse taxels~\cite{integratingpointspreadfunctiontaxelbasedtactilepatternsuperresolution2024}. These examples distinguish the sensor providing supervisory data from the sensor used during deployment.

Electronic-array super-resolution provides related methodological context: the taxel-based SuperTac method~\cite{patel2025supertac} (distinct from the multispectral VBTS of the same name in Section~\ref{subsec:hardware_multimodal}) combines dimensionality reduction with upsampling, while sparse-array super-resolution reconstructs dense pressure patterns from sparse taxels~\cite{kong2025sparsesr}.

Cross-sensor transfer difficulty depends strongly on which hardware boundary is crossed, spanning three progressively harder tiers: (i) \emph{unit-to-unit transfer} to an unseen unit of the same sensor model, which tests tolerance to manufacturing, gel-casting, and lighting variation; (ii) \emph{cross-design optical transfer} across different VBTS families (such as transferring between marker-based, photometric-shading, and curved-fingertip sensors), where both optical encoding and contact mechanics change simultaneously; and (iii) \emph{cross-transducer transfer}, where optical VBTS supervision guides deployment on non-optical magnetic or taxel-based skins. Across all three tiers, target-sensor data used for pretraining, calibration, or fine-tuning contribute directly to adaptation even when unlabeled. Rigorous transfer evaluation therefore needs to specify the held-out hardware tier, the target-sensor data budget, the adapted model components, and the physical task metrics achieved.

Together, shared tactile encoders, signal translation, and cross-transducer supervision target different transfer interfaces with distinct calibration demands. Table~\ref{tab:transfer_methods} compares these cross-sensor transfer mechanisms by the representation or signal transferred, their device-specific interfaces, and the experimental boundaries required to validate generalization.

\begin{table*}[t]
\centering
\caption{Comparison of cross-sensor tactile transfer mechanisms, transferred signals, device-specific interfaces, and evaluation boundaries.}
\label{tab:transfer_methods}
\small
\renewcommand{\arraystretch}{1.12}
\setlength{\tabcolsep}{4pt}
\begin{tabularx}{\textwidth}{@{}P{0.18\textwidth}YYY@{}}
\toprule
\textbf{Representative method} & \textbf{Transferred representation or signal} & \textbf{Device-specific interface} & \textbf{Evidence boundary} \\
\midrule
Sparsh\newline\cite{sparshselfsupervisedtouchrepresentationsvisionbasedtactilesensing2024} & Self-supervised tactile features & Preprocessing and downstream prediction head & Separate pretraining from labeled probe training and target-sensor exposure. \\
T3 / FoTa\newline\cite{transferabletactiletransformersrepresentationlearningdiversesensorstasks2024} & Transformer trunk across sensor--task pairs & Sensor encoder and task decoder & Report zero-/few-shot results by task; transfer is uneven across outputs. \\
AnyTouch / AnyTouch 2\newline\cite{anytouchlearningunifiedstaticdynamicrepresentationmultiplevisuotactilesensors2024,anytouch2generalopticaltactilerepresentationlearningdynamictactileperception2025} & Static and dynamic tactile representations & Sensor observations and alignment data & Separate temporal generalization, object transfer, and physical-device transfer. \\
Touch2Touch\newline\cite{touch2touchcrossmodaltactilegenerationobjectmanipulation2024} & Generated target-domain tactile observations & Trained translation between source and target domains & Test preservation of physical contact and downstream estimates. \\
GenForce\newline\cite{trainingtactilesensorslearnforcesensingeachother2026} & Force knowledge through marker representations & Feature extraction and target calibration/interface & State source and target mechanics, load range, and adaptation effort. \\
SuperMag\newline\cite{supermagvisionbasedtactiledataguidedhighresolutiontactileshapereconstructionmagnetictactilesensors2025} & Optical supervision for magnetic-sensor shape reconstruction & Magnetic input and paired supervisory data & This is cross-transducer learning; the deployed sensor is not a VBTS. \\
\bottomrule
\end{tabularx}
\end{table*}

\subsection{Regime V: Embodied Tactile Intelligence}
\label{subsec:stage_embodied}

The fifth regime focuses on combining tactile observations with other information to support robotic perception, prediction, and action. Datasets may include touch $T$, external vision $V$, proprioception $P$, actions $A$, and language $L$. For action generation, a policy uses the available observation history and an optional instruction:
\begin{equation}
 a_t\sim\pi_{\theta}(\,\cdot\mid o_{\le t}^{V,T,P},\ell\,),
 \label{eq:policy}
\end{equation}
where $a_t$ is the action at time $t$ sampled from the policy $\pi_{\theta}$ parameterized by $\theta$, $o_{\le t}^{V,T,P}$ is the history of external visual ($V$), tactile ($T$), and proprioceptive ($P$) observations up to time $t$, and $\ell$ is an optional language instruction. Future observations can provide training targets, while deployed decisions use observations available at action selection. These data support multimodal alignment, contact prediction, and feedback control.

\subsubsection{\textup{\textbf{Multimodal tactile representations and reasoning}}}
Visual--tactile representation learning combines local contact observations with scene information. Feature sharing, self-supervised fusion, and attention-based recognition learn from both modalities~\cite{vitacfeaturesharingvisiontactilesensingclothtexturerecognition2018,makingsensevisiontouchselfsupervisedlearningmultimodalrepresentationscontactrichtasks2019,vitotransformervisualtactilefusionnetworkobjectrecognition2023,fusiontactilevisualinformationdeeplearningmodelsobjectrecognition2023}. Fused representations support shape and pose estimation by combining tactile evidence at the contact interface with visual observations of the object~\cite{neuralfeelsneuralfieldsvisuotactileperceptioninhandmanipulation2024,inhandobjectposeestimationvisualtactilefusion2025}.

Touch and Go provides human-collected vision and touch observations recorded together~\cite{touchgolearninghumancollectedvisiontouch2022}. Touch100K and TVL add tactile-language descriptions and alignment objectives~\cite{touch100klargescaletouchlanguagevisiondatasettouchcentricmultimodalrepresentation2025,touchvisionlanguagedatasetmultimodalalignment2024}. UniTouch and MMWand connect tactile features with broader multimodal representations~\cite{bindingtoucheverythinglearningunifiedmultimodaltactilerepresentations2024,multimodalrepresentationlearningtactiledata2024}. Octopi and visuotactile-video language models use tactile evidence to describe object properties and reason about interactions~\cite{octopiobjectpropertyreasoninglargetactilelanguagemodels2024,universalvisuotactilevideounderstandingembodiedinteraction2025}. These resources relate contact observations to semantic descriptions, but they do not necessarily include the robot actions needed to learn a manipulation policy.

Integrated hardware also connects perception and action: FingerEye combines continuous proximity/contact feedback with imitation learning, while F-TAC Hand uses distributed optical touch and a training-free grasp-generation algorithm for adaptive control~\cite{fingereyecontinuousunifiedvisiontactilesensingdexterousmanipulation,zhao2025ftachand}.

Recent work also examines how tactile evidence is incorporated into multimodal language models. Touch-R1 uses reinforcement learning with rewards for physical-attribute accuracy and reliance on tactile inputs~\cite{lai2026touchr1}. Splash restricts tactile adaptation to selected language-model parameters to preserve pretrained visual--language capabilities~\cite{park2026splash}.

Multimodal alignment depends on the correspondence between observations. Images and tactile signals from the same object may describe different surface regions, loads, or contact phases. Descriptions such as soft or slippery also depend on the exploratory action. The data can therefore be aligned by object, contact, or time, with different implications for learning. VTDexManip connects visual--tactile pretraining with dexterous manipulation, providing data and task structure for studying how learned representations support control~\cite{vtdexmanipdatasetbenchmarkvisualtactilepretrainingdexterousmanipulationreinforcementlearning2025}.

\subsubsection{\textup{\textbf{Tactile prediction and world models}}}
A tactile world model predicts future observations or contact states from an observation history and candidate actions:
\begin{equation}
 \hat s_{t+1:t+K}=f_{\omega}(s_{\le t},a_{t:t+K-1}),
 \label{eq:world}
\end{equation}
where $f_{\omega}$ is the world model parameterized by $\omega$, $s_{\le t}$ is the history of tactile images or contact features up to time $t$, $a_{t:t+K-1}$ is the sequence of candidate actions from time $t$ to $t+K-1$, $\hat s_{t+1:t+K}$ denotes the predicted future observations or contact states over the next $K$ steps, and $K$ is the prediction horizon. Predicted contact can support planning, generate training trajectories, or provide an expected signal for comparison with subsequent observations. Offline data generation and online control place different demands on prediction speed.

Recent studies investigate predictive tactile models with different representations and control interfaces. Visuo-Tactile World Models and ContactWorld investigate predictive representations and planning, while ViTacWorld emphasizes action-conditioned visuotactile trajectory generation~\cite{visuotactileworldmodels2026a,contactworldwhatrepresentationsmattervisiontactilelatentworldmodelscontactrichmanipulation2026,vitacworldscalingvisuotactileworldmodelscontactrichrobotmanipulation2026}. VT-WAM, Dream-Tac, and $N_0$-TWAM combine future contact prediction with action generation~\cite{vtwamvisualtactileworldactionmodelcontactrichmanipulation2026,dreamtacunifiedtactileworldactionmodelcontactrichrobotmanipulation2026,n0twamscalingtactilenativeworldactionmodel2026}. FeelWorld jointly predicts contact state, a tactile latent representation, and slip state to guide contact-aware planning~\cite{ma2026feelworld}.

Model architecture determines how tactile information influences prediction and action. Tactile-WAM restricts cross-modal attention to preserve the visual prediction path. DexTacWAM represents distributed fingertip contact, while TacWAM uses mechanics-related prediction targets and controls information flow between observations, future supervision, and actions~\cite{tactilewamtouchawareworldactionmodeltactileasymmetricattention2026,dextacwamvisuotactileworldactionmodeldexterousmanipulation2026,tacwamanchorguidedworldactionmodelmechanicsawaretactileprediction2026}. OmniVTA and TacPAC connect anticipated contact with subsequent tactile feedback~\cite{omnivtavisuotactileworldmodelingcontactrichroboticmanipulation2026a,tacpactactilepredictionrealtimeactioncorrectionworldactionmodelscontactrichmanipulation2026}. Differences between predicted and observed contact can then inform action correction, although the discrepancy may also reflect sensor drift or prediction error.

\subsubsection{\textup{\textbf{Tactile-guided policies and control}}}
Multimodal policies use tactile observations alongside other inputs to select actions. GelFusion studies imitation learning under visual constraints, while Tactile ALOHA and simultaneous tactile--visual sensing systems learn manipulation from combined observations~\cite{gelfusionenhancingroboticmanipulationvisualconstraintsvisuotactilefusion2025,tactilealohalearningbimanualmanipulationtactilesensing2025,simultaneoustactilevisualperceptionlearningmultimodalrobotmanipulation2026}. See-Hear-Feel and PolyTouch incorporate additional sensory channels into contact-rich policies~\cite{seehearfeelsmartsensoryfusionroboticmanipulation2022,polytouchrobustmultimodaltactilesensorcontactrichmanipulationusingtactilediffusionpolicies2025}.

T-Rex combines force histories and deformation maps with visuomotor planning. Its variable-rate architecture uses recent tactile feedback to refine actions~\cite{trextactilereactivedexterousmanipulation2026}. Tactile-synchronized robot data support model training, followed by task-specific post-training. OmniVTLA and UniTacVLA connect language and tactile information with action generation, while VTLG addresses task-oriented grasp generation~\cite{omnivtlavisiontactilelanguageactionmodelsemanticalignedtactilesensing2025,unitacvlaunifiedtactileunderstandingpredictionvisionlanguageactionmodels2026,vtlgvisiontactilelanguagegraspgenerationmethodorientedtask2026}. These approaches span different input modalities and control interfaces, ranging from continuous language-conditioned manipulation actions to task-oriented grasp generation.

Other studies explore different interfaces between tactile sensing and pretrained VLAs. VLA-Touch uses tactile-language descriptions for planning and marker-derived feedback for action refinement~\cite{bi2025vlatouch}. TaF-VLA aligns tactile sequences with force measurements before integrating the tactile encoder into a VLA~\cite{huang2026tafvla}. TAP-VLA overlays tactile shear vectors on camera images, allowing existing visual inputs to carry contact information~\cite{vandermerwe2026tapvla}.

The timing of tactile input determines whether a policy can correct an ongoing interaction. A policy that generates several actions at once may use touch only at the beginning of that action sequence. Its ability to react then depends on when new tactile observations enter the next decision. Reactive Diffusion Policy (RDP) addresses this timing mismatch with a slow action-chunk model and a fast tactile-feedback policy~\cite{xue2025reactivediffusion}. AT-VLA adaptively gates tactile input and separates slow visual--language processing from fast tactile action updates~\cite{li2026atvla}. A further failure mode in multimodal policy learning is \emph{modality dominance}: because external vision provides dense, continuous scene cues throughout an episode whereas touch is informative only during brief contact phases, end-to-end policies can learn visual shortcuts and ignore tactile channels unless explicitly regularized or gated. Diagnostic tests that remove, delay, or replace tactile input are therefore crucial for confirming whether a policy genuinely relies on contact feedback to recover from disturbances.

Together, these three lines of work illustrate three complementary architectural interfaces through which touch enters embodied systems: semantic tactile tokens aligned with vision and language for high-level reasoning, action-conditioned latent contact states in world models for predictive planning, and high-frequency local feedback signals in slow--fast visuomotor and VLA policies for closed-loop execution. Table~\ref{tab:embodied_methods} summarizes representative studies across multimodal representations and reasoning, tactile prediction and world models, and tactile-guided policies and control. By comparing each study's training data and sensory modalities against its outputs and functional role, the table highlights how supervision requirements shift from paired vision--touch--language observations for reasoning to action-conditioned histories and tactile-synchronized demonstrations for planning and reactive execution.

\begin{table*}[t]
\centering
\caption{Comparison of representative studies in embodied tactile intelligence across multimodal representation and reasoning, tactile world modeling, and tactile-guided control.}
\label{tab:embodied_methods}
\small
\renewcommand{\arraystretch}{1.12}
\setlength{\tabcolsep}{4pt}
\begin{tabularx}{\textwidth}{@{}P{0.27\textwidth}YY@{}}
\toprule
\textbf{Representative studies} & \textbf{Training data and modalities} & \textbf{Outputs and functional role} \\
\midrule
\multicolumn{3}{@{}l}{\textbf{Multimodal tactile representations and reasoning}} \\
Touch and Go~\cite{touchgolearninghumancollectedvisiontouch2022} & Human-collected visual and tactile observations & Shared visual--tactile representations \\
Touch100K / TVL~\cite{touch100klargescaletouchlanguagevisiondatasettouchcentricmultimodalrepresentation2025,touchvisionlanguagedatasetmultimodalalignment2024} & Tactile observations, visual images, and language descriptions & Alignment of tactile information with visual and language representations \\
Octopi~\cite{octopiobjectpropertyreasoninglargetactilelanguagemodels2024} & Tactile observations and language & Object-property descriptions and interaction reasoning \\
Touch-R1 / Splash~\cite{lai2026touchr1,park2026splash} & Optical tactile and visual observations with language supervision & Physical-property reasoning and tactile descriptions \\
\midrule
\multicolumn{3}{@{}l}{\textbf{Tactile prediction and world models}} \\
ContactWorld / ViTacWorld~\cite{contactworldwhatrepresentationsmattervisiontactilelatentworldmodelscontactrichmanipulation2026,vitacworldscalingvisuotactileworldmodelscontactrichrobotmanipulation2026} & Visuotactile observation histories and candidate actions & Predictive representations and future interaction trajectories for planning \\
FeelWorld~\cite{ma2026feelworld} & Visual observations, 3D tactile point clouds, actions, and contact/slip labels & Joint visual--tactile prediction for contact-aware planning \\
OmniVTA / TacPAC~\cite{omnivtavisuotactileworldmodelingcontactrichroboticmanipulation2026a,tacpactactilepredictionrealtimeactioncorrectionworldactionmodelscontactrichmanipulation2026} & Predicted contact and subsequent tactile observations & Action correction using expected and observed contact \\
\midrule
\multicolumn{3}{@{}l}{\textbf{Tactile-guided policies and control}} \\
GelFusion / Tactile ALOHA~\cite{gelfusionenhancingroboticmanipulationvisualconstraintsvisuotactilefusion2025,tactilealohalearningbimanualmanipulationtactilesensing2025} & Visual and tactile observations with demonstrated actions & Manipulation policies learned from demonstrations \\
RDP / AT-VLA~\cite{xue2025reactivediffusion,li2026atvla} & Demonstrations with visual and tactile/force feedback & Slow--fast policies for tactile correction during action execution \\
T-Rex~\cite{trextactilereactivedexterousmanipulation2026} & Force histories, deformation maps, and visual observations & Action refinement using recent tactile feedback \\
OmniVTLA / UniTacVLA~\cite{omnivtlavisiontactilelanguageactionmodelsemanticalignedtactilesensing2025,unitacvlaunifiedtactileunderstandingpredictionvisionlanguageactionmodels2026} & Vision, touch, language, and action data & Language-conditioned, tactile-informed action generation \\
VLA-Touch~\cite{bi2025vlatouch} & Tactile-language feedback, marker motion, and visual--action data & Semantic replanning and tactile action refinement \\
TaF-VLA~\cite{huang2026tafvla} & Tactile sequences with force labels and manipulation demonstrations & Force-aligned tactile features for VLA control \\
TAP-VLA~\cite{vandermerwe2026tapvla} & Camera images annotated with tactile shear vectors, robot states, and language & VLA actions conditioned on visualized contact cues \\
\bottomrule
\end{tabularx}
\end{table*}

\section{Open Challenges and Future Directions}
\label{sec:challenges}

\begin{figure*}[t]
\centering
\includegraphics[width=\textwidth]{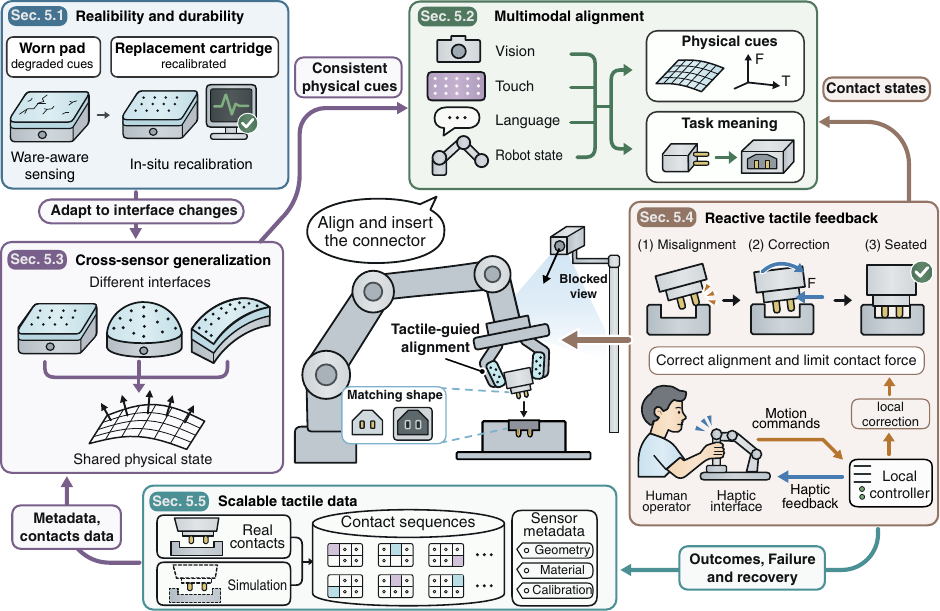}
\caption{System-level integration of the five open challenges in a contact-rich connector alignment and insertion task under visual occlusion: wear-aware, replaceable hardware for durability (Section~\ref{subsec:challenge_durability}); physically grounded multimodal alignment (Section~\ref{subsec:challenge_multimodal}); canonical cross-sensor state mapping (Section~\ref{subsec:challenge_cross_sensor}); low-latency local reactive feedback, optionally coupled with human teleoperation (Section~\ref{subsec:challenge_reactive}); and metadata-rich logging of real and simulated interactions for scalable tactile data (Section~\ref{subsec:challenge_scalable}).}
\label{fig:future_directions_overview}
\end{figure*}

Despite rapid advances in hardware, simulation, and multimodal learning, deploying vision-based tactile sensors reliably in unstructured robotic manipulation remains challenging. Across the physical principles, hardware trade-offs, and data regimes examined in previous sections, five interconnected bottlenecks stand out: physical durability, physically grounded multimodal alignment, cross-sensor generalization, low-latency reactive control, and scalable data reuse. Fig.~\ref{fig:future_directions_overview} illustrates how these five directions connect in a closed-loop manipulation scenario under visual occlusion. For each direction, we examine why the challenge matters, what current methods achieve and leave unresolved, what future research should pursue, and what benefits overcoming it would unlock.

\subsection{Sensor Reliability and Durability}
\label{subsec:challenge_durability}

Long-term deployment in contact-rich manipulation requires sensing interfaces that maintain mechanical and optical integrity under repeated physical stress. Unlike remote cameras, VBTS elastomers directly endure cyclic indentation, high shear, sharp-edge tearing, and abrasive sliding. Over time, surface wear, reflective-coating delamination, optical hazing, marker distortion, and viscoelastic creep alter both contact mechanics and image appearance, silently invalidating calibration and degrading force, shape, and slip estimates. Current designs mitigate maintenance costs through protective outer skins and modular replaceable cartridges, as in PolyTouch~\cite{polytouchrobustmultimodaltactilesensorcontactrichmanipulationusingtactilediffusionpolicies2025}, while self-healing liquid crystalline gels suggest a route toward autonomous, skin-like recovery from cuts and punctures~\cite{lin2026selfhealingliquidcrystalline}. However, replacing worn pads often introduces mounting tolerances and optical shifts that require recalibration, and self-healing formulations must still match conventional elastomers in optical clarity, low viscoelastic hysteresis, and rapid mechanical recovery.

Future work should co-design abrasion-resistant coatings, low-hysteresis elastomers, and mechanically keyed replaceable pads that preserve optical and geometric registration across swaps. On the software side, sensors should incorporate online self-monitoring to detect coating wear, illumination drift, or stiffness changes and trigger lightweight, in-situ recalibration. Standardized endurance benchmarks are also needed to track mechanical response, optical cue fidelity, and force or shape accuracy across thousands of shear, impact, replacement, and healing cycles. Achieving durable, self-monitoring VBTSs would enable continuous, unattended operation in industrial assembly, logistics, and household robotics, reducing maintenance downtime and preventing silent calibration drift from causing manipulation failures.

\subsection{Multimodal Alignment}
\label{subsec:challenge_multimodal}

Embodied reasoning and dexterous manipulation rarely rely on touch alone; robots must link local contact events with external visual context, proprioception, language instructions, and motor actions. Aligning these modalities is fundamentally challenging because they differ sharply in spatial scale (global scene views versus millimeter-scale occluded contact patches), temporal dynamics (brief, high-frequency contact transients versus slower visual and language streams), and abstraction level (continuous physical quantities such as shear, compliance, and incipient slip versus semantic descriptions that depend on the robot's exploratory motion). Current approaches have begun bridging these gaps: TVL aligns tactile observations with vision and language in a shared representation space, while UniTacVLA integrates tactile understanding and contact prediction directly with language-conditioned action generation~\cite{touchvisionlanguagedatasetmultimodalalignment2024,unitacvlaunifiedtactileunderstandingpredictionvisionlanguageactionmodels2026}. Yet compressing tactile images into coarse, language-aligned embeddings often washes out fine geometric gradients and force cues needed for contact-rich control, and multimodal policies can still over-rely on external vision when touch is informative only during brief contact phases.

Future work should develop dual-granularity or physically grounded alignment architectures that synchronize tactile streams with robot kinematics and retain quantitative contact states (force, shear, local geometry, and slip) alongside semantic features. Training objectives should explicitly enforce tactile grounding rather than allowing vision to dominate, and evaluation should include controlled diagnostic tests---such as visual occlusion or visually identical objects with distinct friction, texture, or compliance---to verify that aligned representations drive both reasoning and control. Achieving faithful multimodal alignment would allow embodied agents to connect high-level semantic instructions (such as handling fragile or slippery objects) directly to precise physical execution, sustaining reliable manipulation even when visual cues are occluded or ambiguous.

\subsection{Cross-Sensor Generalization}
\label{subsec:challenge_cross_sensor}

Although Section~\ref{subsec:stage_cross_sensor} reviewed early advances in shared tactile encoders, cross-domain image translation~\cite{touch2touchcrossmodaltactilegenerationobjectmanipulation2024}, and marker-based force transfer~\cite{trainingtactilesensorslearnforcesensingeachother2026}, true plug-and-play deployment across changing hardware remains out of reach. Most existing methods validate transfer on offline perception benchmarks or narrow source--target pairs, leaving closed-loop manipulation policies and multimodal models fragile whenever physical hardware changes. In real robotic systems, cross-sensor shift is not merely a one-time dataset mismatch: a single dexterous hand often combines flat, domed, and curved sensing interfaces, while routine operation introduces wear-induced drift and cartridge replacements (Section~\ref{subsec:challenge_durability}). Under these conditions, matching raw image appearance or semantic embeddings is insufficient; downstream reasoning and reactive control fail unless different interfaces produce metrically consistent contact geometry, traction, and compliance cues without costly policy retraining.

Future work should shift from device-bound pixel features toward canonical, physics-grounded intermediate states---such as calibrated 3D surface deformations, contact-patch geometry, and normalized traction fields (Fig.~\ref{fig:future_directions_overview})---that isolate upstream multimodal models and controllers from sensor-specific optics and form factors. Conditioning these canonical mappers on compact sensor metadata (geometry, elastomer properties, and calibration parameters from Section~\ref{subsec:challenge_scalable}) alongside brief, self-supervised exploratory touches could enable rapid, automated adaptation after pad swaps or hand upgrades. Evaluation should likewise move beyond offline feature probes toward closed-loop hardware-migration benchmarks that quantify manipulation success, force-control fidelity, and adaptation effort when deploying the same embodied policy across replacement cartridges, unseen fingertip curvatures, and hybrid sensor suites. Achieving hardware-agnostic transfer at the system level would finally decouple tactile intelligence from individual sensor assemblies, allowing pretrained perception models and manipulation policies to persist seamlessly across sensor replacements, hardware generations, and diverse robot morphologies.

\subsection{Reactive Tactile Feedback}
\label{subsec:challenge_reactive}

In contact-rich tasks such as fragile grasping, precision insertion, and in-hand reorientation, critical events including incipient slip, edge collision, and force overshoot unfold within tens of milliseconds. Tactile feedback is therefore effective only if new contact observations modulate robot actions before a grasp fails or an object is damaged. In practice, end-to-end latency accumulates across elastomer viscoelastic recovery, camera exposure and readout, tactile feature extraction, and policy inference, creating a timing mismatch between millisecond-scale contact dynamics and slower action-chunking or VLA policies. Current slow--fast architectures such as Reactive Diffusion Policy and T-Rex address this gap by pairing lower-frequency visuomotor planning with faster tactile action refinement~\cite{xue2025reactivediffusion,trextactilereactivedexterousmanipulation2026}. Even so, determining when local contact changes should trigger immediate overrides---and ensuring that rapid corrections remain dynamically stable without oscillating or diverging from task-level plans---remains an open problem.

Future work should co-design fast-recovering elastomers and high-frame-rate or event-based imaging with lightweight, low-latency tactile encoders and event-triggered reactive controllers that adaptively gate local corrections into higher-level plans. Beyond overall task success, closed-loop benchmarks should explicitly measure end-to-end sensing-to-actuation delay, peak contact-force overshoot, incipient-slip arrest rate, and recovery under unmodeled external disturbances. Achieving low-latency reactive tactile feedback would give robots reflex-like dexterity, enabling them to arrest slip before dropping objects, regulate contact forces safely on delicate or unfamiliar geometries, and recover smoothly from unexpected physical perturbations in real time.

\subsection{Scalable Tactile Data}
\label{subsec:challenge_scalable}

General-purpose tactile intelligence requires large, diverse datasets spanning varied objects, contact geometries, and manipulation tasks, yet physical data collection is slow, wear-prone, and hardware-specific. Existing tactile corpora remain fragmented across incompatible sensors, illumination setups, sampling rates, and collection protocols. Unlike web-scale vision datasets, raw tactile images rarely record the underlying sensor geometry, elastomer stiffness, camera calibration, or physical force scale needed to interpret contact consistently across laboratories, limiting cross-dataset reuse and fair comparison. Current efforts such as T3 aggregate heterogeneous multi-sensor datasets into a unified training format, while Sparsh uses self-supervised pretraining to learn reusable features without dense task-specific labels~\cite{transferabletactiletransformersrepresentationlearningdiversesensorstasks2024,sparshselfsupervisedtouchrepresentationsvisionbasedtactilesensing2024}. Nevertheless, pooling raw frames without standardized metadata, quality filtering, or physical normalization can over-represent redundant static contacts, conflate incompatible counting units, and obscure true contact diversity.

Future work should establish open tactile data schemas that bundle observation streams with standardized hardware descriptors, calibration parameters, sampling rates, and explicit counting units. Reusable software pipelines are likewise needed for cross-dataset normalization, automated contact-event filtering, cross-modal auto-annotation (leveraging synchronized force--torque transducers or robot kinematics), and physics-consistent simulation or generative augmentation. Progress should be evaluated by non-redundant contact diversity and downstream sample efficiency rather than raw frame volume alone. Realizing scalable, metadata-rich tactile corpora and curation pipelines would allow multi-institution datasets to support foundation-scale pretraining, drastically reduce repeated data collection for new tasks, and establish reproducible scaling laws for tactile learning.

\section{Conclusion}
\label{sec:conclusion}

This review has examined vision-based tactile sensors (VBTSs) as integrated sensing-and-learning systems spanning contact mechanics, hardware co-design, data regimes, and embodied robotic manipulation. Examining optical transduction alongside classical contact-mechanics and photometric models shows why idealized theory alone cannot deliver robust real-world touch perception: finite elastomer geometry, nonlinear viscoelasticity, ill-posed inverse mappings, and optical calibration sensitivity couple physical sensor design directly to downstream estimation. Organizing hardware advances around six competing goals---miniaturization, large-area coverage, spatial resolution, force measurement performance, temporal performance, and multimodal sensing---demonstrates that every physical choice across elastomers, illumination, optics, and cameras reshapes calibration overhead and computational demands. Mapping perception, simulation, transfer, and embodied policies into five data and learning regimes further clarifies how supervision bottlenecks, unit-explicit data budgets, and generalization boundaries evolve from single-sensor physical measurement to tactile world models and vision-tactile-language-action control.

Taken together, these perspectives yield three practical takeaways for advancing tactile robotics. First, hardware and algorithms should be engineered as a closed co-design loop: sensor morphology, elastomer compliance, and optical encoding should be selected for the physical observables and temporal bandwidth required by the target manipulation behavior, while calibration residuals and closed-loop failure modes should feed back into physical redesign. Second, progress in tactile learning depends as much on representation choice and data hygiene as on raw dataset scale; choosing between calibrated physical states, low-dimensional contact abstractions, and multimodal embeddings---while reporting simulated and physical budgets in explicit, compatible counting units---is essential for reproducible scaling and transfer. Third, evaluation must move beyond offline visual fidelity toward closed-loop physical rigor, using controlled occlusion, disturbance, and hardware-migration tests to verify that tactile signals genuinely govern contact-rich execution rather than being overshadowed by external vision.

Looking ahead, turning VBTSs from bespoke laboratory prototypes into durable, reusable components of general-purpose embodied robots will require coordinated progress across the five interconnected priorities synthesized in Fig.~\ref{fig:future_directions_overview}: sensor reliability and durability, multimodal alignment, cross-sensor generalization, reactive tactile feedback, and scalable tactile data. Closing this loop---combining wear-aware, replaceable sensing interfaces with physically grounded multimodal representations, hardware-agnostic state mappers, low-latency local control, and metadata-rich contact corpora---offers a clear path toward embodied robotic systems that perceive and manipulate the physical world with human-like dexterity and resilience.

\section*{Acknowledgements}

The authors used Google Gemini and OpenAI GPT to edit the manuscript's language and formatting. Because AI-generated edits can introduce errors or alter technical meaning, the authors take responsibility for the accuracy of the final text and citations.

\section*{Author Contributions}

\textbf{Peng Zhou} conceived the review framework, coordinated the project, and co-led drafting and revision. \textbf{Jun Hu} co-led drafting, audited the data counts, prepared the figures and tables, and developed the companion website. \textbf{Sihan Chen} curated the literature and co-drafted the hardware, sensing mechanics, and manipulation sections. \textbf{Zeqing Zhang} contributed analysis and writing on force sensing, multimodal perception, and deformable-object manipulation. \textbf{Haofei Ma} synthesized the literature and co-drafted the sections on simulation, datasets, and manipulation policies.

\textbf{Zhenyu Lu} reviewed sensor hardware, teleoperation, and tactile feedback. \textbf{Sichao Liu} synthesized and reviewed work on tactile world models, multimodal foundation models, and industrial manipulation. \textbf{Xueqian Wang} provided technical guidance and reviewed tactile feedback, teleoperation, and embodied manipulation. \textbf{Pai Zheng} reviewed industrial deployment, sensor durability, and scalable tactile data. \textbf{Xiang Li} reviewed dexterous manipulation and reactive tactile control. \textbf{Shan Luo} reviewed tactile perception, cross-sensor transfer, and visuo-tactile learning. \textbf{Jia Pan} reviewed sensing mechanics, cross-transducer transfer, and sim-to-real benchmarks. \textbf{David Navarro-Alarcon} reviewed contact mechanics, sensor-based control, and multimodal alignment. \textbf{Chenguang Yang} reviewed learning from demonstration, haptic feedback, and embodied intelligence. \textbf{Michael Yu Wang} helped frame the review and critically revised the manuscript.

\section*{Statements and Declarations}

\subsection*{Ethical considerations}
Not applicable.

\subsection*{Consent to participate}
Not applicable.

\subsection*{Consent for publication}
Not applicable.

\subsection*{Declaration of conflicting interest}
The authors declared no potential conflicts of interest with respect to the research, authorship, and/or publication of this article.

\subsection*{Funding statement}
This work was supported by the Basic and Applied Basic Research Foundation of Guangdong Province, the National Natural Science Foundation of China (grant no.~62403211), and the Youth S\&T Talent Support Programme of the Guangdong Provincial Association for Science and Technology (award no.~SKXRC2025092).

\subsection*{Data availability}
The review uses data and results reported in the cited literature. The study-level counts and conversion notes used to compare reported data budgets in Fig.~\ref{fig:five_stage_roadmap} can be explored on the \href{https://gmi-lab-2025.github.io/VBTSwebsite/}{companion website}.

\bibliographystyle{SageH}
\bibliography{ijrr_references}
\end{document}